\pdfoutput=1
\documentclass[final,5p,times,twocolumn]{elsarticle}

\biboptions{numbers,sort&compress}

\usepackage{amsmath,amssymb,amsthm}
\usepackage{booktabs}
\usepackage{graphicx}
\usepackage{xcolor}

\usepackage[colorlinks=true,linkcolor=blue!60!black,citecolor=blue!60!black,urlcolor=blue!60!black,pdftitle={Disentangling Monitor Hardware Artifacts and Smoothing Lag for Shift-Robust Intraoperative EEG and Hemodynamic Risk Calibration},pdfauthor={Zeyu Jia},pdfcreator={LaTeX with elsarticle},pdfproducer={pdfTeX}]{hyperref}

\graphicspath{{./}{figures/}}
\newcommand{\inputtable}[1]{\IfFileExists{tables/#1}{\input{tables/#1}}{\input{#1}}}

\newtheorem{proposition}{Proposition}

\journal{arXiv}
\myfooter[L]{Preprint}

\begin{document}

\begin{frontmatter}

\title{Disentangling Monitor Hardware Artifacts and Smoothing Lag for Shift-Robust Intraoperative EEG and Hemodynamic Risk Calibration}

\author[tmu,tju]{Zeyu Jia\corref{cor1}}
\ead{jiazeyu@tju.edu.cn}
\cortext[cor1]{Corresponding author.}

\affiliation[tmu]{organization={School of Biomedical Engineering and Technology, Tianjin Medical University},
            addressline={No. 22 Qixiangtai Road, Heping District},
            city={Tianjin},
            postcode={300070},
            country={China}}
\affiliation[tju]{organization={Medical School, Tianjin University},
            addressline={No. 92 Weijin Road, Nankai District},
            city={Tianjin},
            postcode={300072},
            country={China}}

\begin{abstract}
\textbf{Background and Objective:} Commercial intraoperative electroencephalography (EEG) and blood pressure monitors introduce cyclic impedance pulses, smoothing delays, direct-current (DC) drift, and cuff holds that confound risk forecasting and uncertainty calibration. We developed a causal signal-conditioning and shift-robust conformal calibration framework across $5,439$ public anesthesia procedures.

\textbf{Methods:} Streaming filters blank $10.13\text{-min}$ $64\text{-Hz}$ impedance checks, apply unclipped post-notch conditioning, separate $63\text{-s}$ suppression-ratio lag, and distinguish three-channel cuff inflations from dropouts across propofol--remifentanil total intravenous anesthesia (TIVA; $N = 3,214$), volatile anesthesia ($N = 2,046$), European endoscopy sedation ($N = 171$), and DC-coupled surgery ($N = 8$). De-impedanced cortical tree ensembles were compared against deep sequence models, paired with patient-cluster Subgroup-Covariate-Weighted Conformal Risk Control (CRC) and lag-buffered online Conformal Proportional--Integral--Derivative (PID) tracking.

\textbf{Results:} Pre-filter blanking removed $97.90\%$ and $97.60\%$ of false flatline conflicts while preserving $99.23\%$ and $98.96\%$ of deep suppression. On unseen TIVA surgeries ($N = 525$), cortical features achieved area under the precision--recall curve (AUPRC) $0.3135$ ($1.94\%$ prevalence) and $0.1627$ ($0.82\%$ lag-free prevalence), with no detectable difference from deep encoders ($p \ge 0.36$) and no cortical gain for hypotension ($0.3432$ vs.\ $0.3437$, $p = 0.784$). Zero-retraining volatile transfer ($N = 1,048$) showed no detectable difference from an end-tidal gas oracle on lag-free suppression ($0.0820$ vs.\ $0.0736$, $p = 0.472$). Subgroup-Covariate-Weighted CRC bounded all $70$ TIVA shift cells at $\le 4.38\%$ ($\alpha = 5.0\%$) in $66.32\,\text{ms}$ per $30\text{-s}$ epoch.

\textbf{Conclusions:} Causal artifact disentangling and shift-aware conformal calibration support reliable real-time intraoperative risk bounds within explicit physiological observability limits.
\end{abstract}


\end{frontmatter}
\thispagestyle{pprintTitle}

\section{Introduction}
\label{sec:intro}

Intraoperative burst suppression on frontal electroencephalography (EEG) and mean arterial pressure (MAP) excursions below $65\,\text{mmHg}$ represent actionable physiological hazards during general anesthesia and procedural sedation \cite{L04_Purdon2015,L_Wildes2019,L_Wijnberge2020}. Although machine learning models increasingly forecast anesthetic depth and intraoperative hypotension from perioperative waveforms \cite{L_Lee2018,L_Hatib2018,L35_Lee2021,L43_Sun2026}, bedside telemetry rarely delivers clean cortical or vascular measurements directly. Instead, clinical archives record composite streams shaped by monitor hardware diagnostics, proprietary smoothing windows, and intermittent cuff broadcast protocols \cite{L33_Rampil1998,L_Pilge2006,L44_Wax2011}.

Four measurement artifacts systematically distort naïve modeling on bedside recordings. First, commercial Bispectral Index (BIS) monitors inject periodic $64\text{-Hz}$ electrode-impedance test currents approximately every $10.13\,\text{min}$ \cite{L27_Lee2022}. If raw $128\text{-Hz}$ EEG is passed directly into a $[0.5, 30]\,\text{Hz}$ bandpass filter without pre-filter rail and Nyquist-carrier blanking, step recovery and Nyquist attenuation produce multi-second $< 5\,\mu\text{V}$ flatlines while the monitor's internal Suppression Ratio (SR) correctly reads zero. Second, external hospital archives recorded on unnotched or direct-current (DC)-coupled amplifiers exhibit $50/60\text{-Hz}$ mains hum and millivolt $\text{Ag}/\text{AgCl}$ baseline offsets \cite{L53_Garbe2026_DOSE1,L54_SainiFerron2023_ds004541}; applying fixed microvolt amplitude gates or pre-bandpass hard clipping before mains removal wipes out valid cortical epochs. Third, commercial BIS-SR averages isoelectricity over a trailing $63\text{-s}$ boxcar \cite{L33_Rampil1998}, conflating genuine $5\text{-min}$ lookahead warning with post-suppression trailing holdover. Fourth, oscillometric noninvasive blood pressure (NIBP) cuffs broadcast held numerical values every $2\,\text{s}$ between physical inflations \cite{L44_Wax2011,L45_Maheshwari2018}, confounding true pressure arrivals with telemetry dropouts and inflating apparent sample counts.

Even after isolating authentic physiological events, point risk scores remain insufficient for clinical decision support under perioperative distribution shifts \cite{L21_Ionescu2026,L_Seoni2026_CMPB,L_Fayyad2024_CMPB}. Calibration sets collected during stable propofol--remifentanil total intravenous anesthesia (TIVA) maintenance face three simultaneous deployment shifts: (i) non-stationary within-surgery transitions across bolus induction and emergence, (ii) cross-anesthetic transfer to sevoflurane and desflurane volatile maintenance where target-controlled infusion (TCI) pump trajectories are absent and frontal theta--alpha spectra shift \cite{L52_Akeju2014}, and (iii) demographic-and-modality prevalence shifts toward older adults monitored solely via intermittent NIBP cuffs \cite{L04_Purdon2015,L44_Wax2011}. Pooling epoch-level residuals across surgeries of unequal duration violates patient-level exchangeability \cite{L_Dunn2023,L_LeeBarber2026}, exposing geriatric cortical and cuff-monitored subgroups to systematic safety-bound under-coverage.

\begin{figure*}[!t]
\centering
\includegraphics[width=0.92\textwidth]{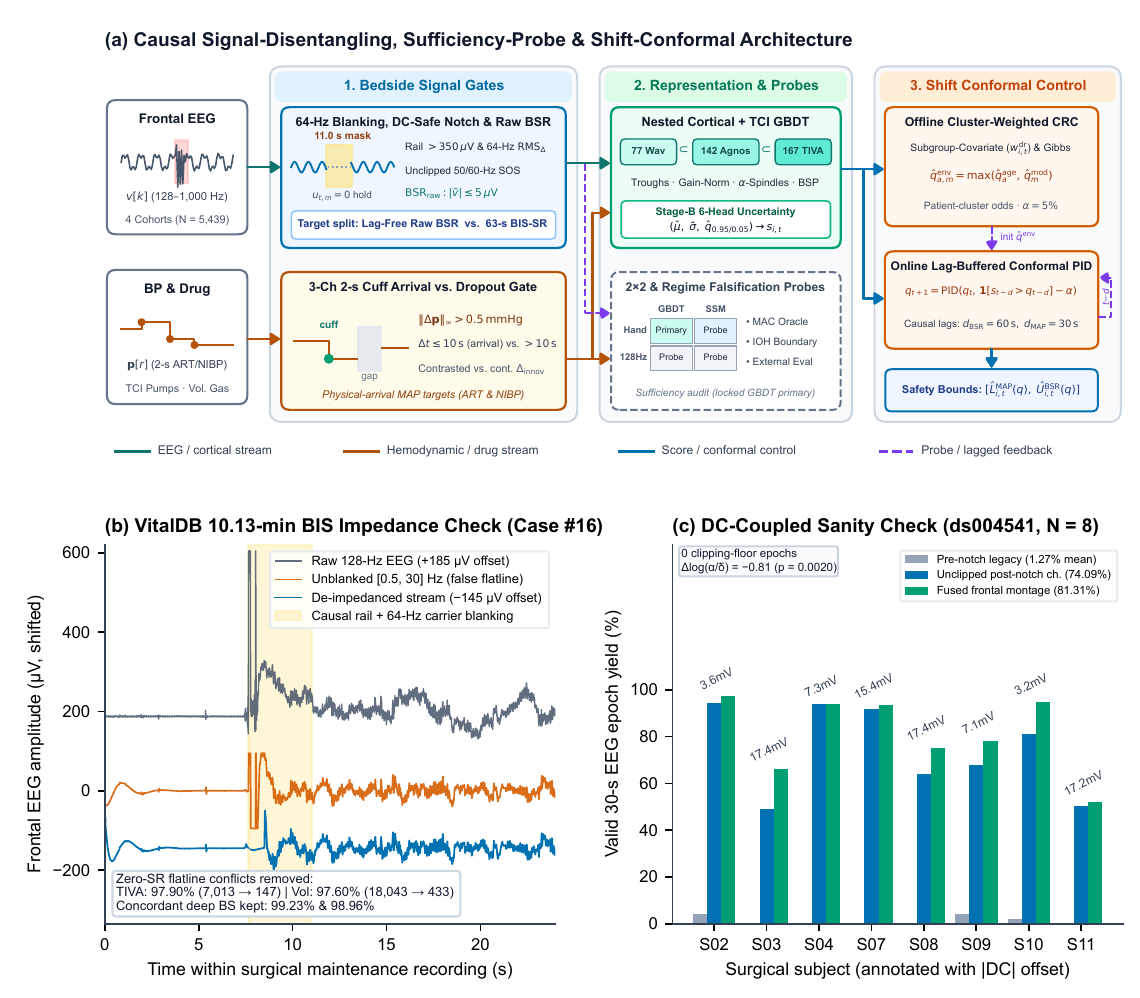}
\caption{\textbf{Three-pillar streaming architecture and bedside hardware artifact disentangling across $5,439$ public procedures.} (a) Causal conditioning, representation-sufficiency probes, and three-shift conformal risk control. (b) Real $128\text{-Hz}$ VitalDB maintenance trace showing how pre-filter rail and $64\text{-Hz}$ carrier blanking eliminates post-filter false flatlines while preserving genuine deep burst suppression. (c) Qualitative DC-coupled surgical EEG check on OpenNeuro ds004541 ($N = 8$) demonstrating recovery of valid frontal epochs across $3.2$--$17.4\,\text{mV}$ baseline offsets without clipping-floor distortion.}
\label{fig:dsp}
\end{figure*}

To address these coupled signal-processing and statistical challenges, we develop and validate a three-pillar streaming framework across $5,439$ quality-controlled (QC-valid) procedures from four public cohorts in South Korea, Europe, and South America (Fig.~\ref{fig:dsp}, Table~\ref{tab:cohort}). Our primary contributions are as follows:
\begin{itemize}
    \item \textbf{Causal bedside signal and hardware artifact disentangling:} We formulate streaming pre-filter blanking of $10.13\text{-min}$ $64\text{-Hz}$ impedance checks, unclipped DC-safe post-notch conditioning for external AC/DC-coupled monitors, lag-free versus $63\text{-s}$ smoothed suppression separation, and native $2\text{-s}$ three-channel cuff inflation versus $>10\text{-s}$ dropout accounting.
    \item \textbf{Full-registry representation sufficiency and physiological boundary mapping:} Using controlled $2\times 2$ cross-plugging across $N = 3,214$ TIVA and $N = 2,046$ volatile surgeries plus two external cohorts, we show that de-impedanced cortical trough and $\alpha$-spindle features in gradient-boosted decision trees (GBDT) outperform clinical monitor and state-space baselines with no detectable difference from $128\text{-Hz}$ causal ConvNeXt-1D and selective state-space sequence models on lag-free suppression ($p \ge 0.36$), and we delineate clear negative boundaries where cortical EEG or discrete cuff steps confer no incremental lift.
    \item \textbf{Shift-robust subgroup-covariate and online conformal risk control:} We establish patient-cluster Subgroup-Covariate-Weighted Conformal Risk Control (CRC), Overlapping-Group CRC, and lag-buffered Causal Online Conformal Proportional--Integral--Derivative (PID) calibration across phase, anesthetic-regimen, and ward-prevalence shifts at a bedside latency of $66.32\,\text{ms}$ per $30\text{-s}$ epoch.
\end{itemize}

\begin{table*}[!t]
\centering
\caption{Multi-cohort perioperative scope and bedside hardware artifact disentangling across 5,439 QC-valid public anesthesia procedures.}
\label{tab:cohort}
\footnotesize
\setlength{\tabcolsep}{1.8pt}
\begin{tabular*}{\textwidth}{@{\extracolsep{\fill}}lrrllll}
\toprule
\textbf{Cohort \& regimen} & \textbf{Cases} & \textbf{Epochs} & \textbf{Hardware} & \textbf{Artifact} & \textbf{Mitigation or yield} & \textbf{Preservation or role} \\
\midrule
VitalDB TIVA & $3,214$ & $1,151,353$ & $128\text{-Hz}$ BIS + BP & $64\text{-Hz}$ \& $63\text{-s}$ lag & Train $1,834$; test $525$ & Scaling \& Shift-1 CRC \\
\quad Signal-audit set & $500$ & $175,766$ & $128\text{-Hz}$ EEG + NIBP & $8,628$ checks ($10.1\text{-m}$) & $97.90\%$ conflict cut & $99.23\%$ deep BS ($4,001/4,032$) \\
\quad Cuff $2\text{-s}$ audit & $455$ & $1,294,016^\dagger$ & $3\text{-ch}$ NIBP cuff & $98.85\%$ held repeats & $14,019$ infl.\ vs.\ $825$ drop. & ${+}5.98\%$ pulse-press.\ only \\
VitalDB Volatile & $2,046$ & $760,986$ & $128\text{-Hz}$ BIS + Primus & $11,739$ checks ($10.1\text{-m}$) & $97.60\%$ conflict cut & $98.96\%$ deep BS ($N{=}1,048$) \\
DOSE-I sedation & $171$ & $9,348$ & $125\text{-Hz}$ Intellivue & $50\text{-Hz}$ \& $83.24\%$ holds & $97.06\%$ valid ($0$ wipeout) & $1,440$ boluses; directional \\
ds004541 surgery & $8$ & $672$ & $1\text{-kHz}$ DC EDF+C & $3.2$--$17.4\,\text{mV}$ drift & $74.09\%$ yield ($1.27\%$ raw) & Qualitative sanity ($0$ floor) \\
\midrule
Total QC-valid & $\mathbf{5,439}$ & $1,922,359$ & $3$ countries, $4$ cohorts & Check, DC, lag, cuff & Causal conditioning & $0$ overlap across $11$ splits \\
\bottomrule
\end{tabular*}
\par\vspace{2pt}\noindent{\footnotesize Notes: TIVA, propofol--remifentanil total intravenous anesthesia; BS, burst suppression; SR, commercial $63\text{-s}$ suppression ratio; NIBP, oscillometric noninvasive blood pressure; CRC, Conformal Risk Control. The $500$-case signal-audit subset ($455$ with NIBP; $^\dagger$native $2\text{-s}$ packets) is nested within the $3,214$-case TIVA registry and excluded from the Total row sum.}
\end{table*}

\section{Related Work}
\label{sec:related}

\subsection{Intraoperative EEG dynamics and hemodynamic early warning}
Age-dependent frontal $\alpha$-band ($8$--$14\,\text{Hz}$) attenuation \cite{L04_Purdon2015}, aperiodic $1/f$ spectral steepening \cite{L01_Donoghue2020,L02_Soehle2026}, and transient $\alpha$-spindle collapses \cite{L42_Cartailler2019,L43_Sun2026} precede metabolic burst suppression \cite{L28_Ching2012}. State-space Burst Suppression Probability (BSP) filters \cite{L34_Chemali2013} mitigate the $63\text{-s}$ boxcar delay of commercial BIS-SR \cite{L33_Rampil1998,L_Pilge2006}, yet remain vulnerable to unblanked impedance-check pulses in open surgical archives \cite{L27_Lee2022}. In parallel, arterial waveform and multi-vital hypotension predictors \cite{L_Hatib2018,L35_Lee2021,L_Pal2024_CMPB} require careful separation from selection bias \cite{L_Enevoldsen2022} and intermittent oscillometric cuff step bias \cite{L44_Wax2011,L45_Maheshwari2018}. To isolate where model complexity confers genuine gain, we cross-plug representation design (front-end waveform features vs.\ $128\text{-Hz}$ convolutions \cite{L_Liu2022_ConvNeXt}) with temporal aggregation (back-end GBDT \cite{L46_Grinsztajn2022} vs.\ selective diagonal state-space models \cite{L22_Gu2022,L29_Gu2024,L38_Dao2024,L39_Tallec2018}).

\subsection{Conformal uncertainty calibration under clinical shift}
Recent biomedical studies emphasize separating diagnostic sensitivity and uncertainty calibration objectives from raw point-prediction accuracy across physiological waveform and clinical imaging models \cite{L_Seoni2026_CMPB,L_Bomberna2024_CMPB,L_Fayyad2024_CMPB}. Standard split conformal prediction and Conformalized Quantile Regression (CQR) \cite{L_Lei2018,L_Romano2019} extend to monotone bounded losses via Conformal Risk Control \cite{L23_Angelopoulos2024}, hierarchical repeated measurements via cluster inverse weighting \cite{L_Dunn2023,L_LeeBarber2026}, covariate shift via likelihood-ratio reweighting \cite{L55_Tibshirani2019}, overlapping subgroups via conditional quantile regression \cite{L32_Gibbs2025}, and non-stationary time series via Adaptive Conformal Inference and Conformal PID control \cite{L24_Gibbs2021,L_XuXie2021,L30_Angelopoulos2023,L31_Barber2023}. Here we unify patient-cluster subgroup conditioning, density-ratio covariate reweighting, and lag-buffered online residual tracking across intraoperative phase, drug, and ward-composition shifts.

\section{Materials and Methods}
\label{sec:methods}

\subsection{Multi-cohort datasets and zero-peek subject-disjoint protocol}
We analyzed $5,439$ QC-valid procedures ($1,922,359$ non-overlapping $30\text{-s}$ epochs) across four public repositories (Table~\ref{tab:cohort}; complete screening funnels in Supplementary Table~S1):
\begin{enumerate}
    \item \textbf{VitalDB TIVA registry ($N = 3,214$):} From $6,388$ non-cardiac surgeries at Seoul National University Hospital \cite{L27_Lee2022}, deterministic screening of propofol--remifentanil TIVA recordings with synchronized $128\text{-Hz}$ raw frontal EEG, BIS telemetry, and $2\text{-s}$ invasive radial arterial (ART) or cuff (NIBP) pressure yielded $3,215$ first-surgery deduplicated subjects, of whom $3,214$ ($99.97\%$, $1,151,353$ epochs, $9,594.6\,\text{h}$) met the $\ge 20$-epoch ($10\text{-min}$) duration criterion. Deterministic SHA-256 stratification by age ($\ge 65$ vs.\ $< 65\,\text{yr}$) and pressure modality allocated $1,834$ training (nesting a $220$-case signal-audit subsample alongside $1,614$ remaining training subjects), $260$ validation, $315$ calibration, $525$ locked unseen test ($185,162$ epochs), and $280$ disjoint signal-audit holdout procedures ($500$ total signal-audit cases, $455$ with NIBP).
    \item \textbf{VitalDB Volatile registry ($N = 2,046$):} Screening volatile surgeries (Primus sevoflurane or desflurane) with strictly zero propofol infusion and zero TIVA subject overlap yielded $2,046 / 2,050$ ($99.80\%$, $760,986$ epochs, $6,341.6\,\text{h}$) QC-valid procedures split into $649$ training, $349$ validation/calibration, and $1,048$ locked test surgeries ($393,024$ epochs: $839$ sevoflurane, $365$ desflurane, including $156$ co-administered cases).
    \item \textbf{DOSE-I European endoscopy sedation ($N = 171$):} All $171$ open-access gastrointestinal and bronchoscopic propofol sedation recordings \cite{L53_Garbe2026_DOSE1} (Philips Intellivue, two-channel $125\text{-Hz}$ EEG, NIBP, $1,440$ discrete $10\text{-mg}$ propofol boluses, and Modified Observer's Assessment of Alertness/Sedation [MOAA/S] scores; $9,348$ epochs, $77.9\,\text{h}$) were evaluated strictly externally without inputting benchmark artifact annotations into model inference.
    \item \textbf{OpenNeuro ds004541 surgical general anesthesia ($N = 8$):} Eight $1,000\text{-Hz}$ DC-coupled frontal surgical EEG recordings \cite{L54_SainiFerron2023_ds004541} ($672$ total $30\text{-s}$ epochs across awake baseline, pre-loss-of-consciousness induction, surgical maintenance, pre-recovery emergence, and post-recovery wakefulness) served strictly as a qualitative physiological and millivolt DC-drift conditioning sanity check.
\end{enumerate}
Pairwise subject-identifier overlap across all $11$ VitalDB splits was verified to be zero. All feature normalizers, out-of-fold neural stems, probability calibrators, and conformal cutoffs were frozen prior to single-shot locked-test evaluation.

\subsection{Causal bedside signal conditioning and endpoint disentangling}
\label{sec:dsp}

Let $v[k]$ denote raw frontal EEG sampled at rate $f_s$, partitioned into causal $0.5\text{-s}$ sub-windows $\mathcal{S}_m$ ($m = 1, \dots, 60$ per $30\text{-s}$ epoch $t$, initial sample $k_0$). Commercial $128\text{-Hz}$ BIS Vista streams inject cyclic impedance checks at the Nyquist frequency ($64\,\text{Hz}$, alternating sign $(-1)^k$). Before IIR filtering, our BIS Vista front-end flags $\mathcal{S}_m$ as rail artifact whenever $\max_{k \in \mathcal{S}_m}|v[k]| > 350\,\mu\text{V}$, intra-sub-window slew $\max_{k \in \mathcal{S}_m \setminus \{k_0\}}|v[k]-v[k-1]| > 200\,\mu\text{V}$, or inter-sub-window jump $|v[k_0]-v[k_0-1]| > 200\,\mu\text{V}$, and flags $64\text{-Hz}$ carrier dominance whenever
\begin{equation}
\label{eq:blank}
\begin{aligned}
\text{RMS}_{\Delta}(\mathcal{S}_m) &= \sqrt{\frac{1}{|\mathcal{S}_m|-1}\sum_{k \in \mathcal{S}_m \setminus \{k_0\}}\bigl(v[k]-v[k-1]\bigr)^2} \\
&> \max\bigl(1.2\,\mu\text{V},\, 3.5\max(\sigma_{\text{bp}}(\mathcal{S}_m),\, 0.25\,\mu\text{V})\bigr),
\end{aligned}
\end{equation}
where $\sigma_{\text{bp}}(\mathcal{S}_m)$ is the $[0.5, 30]\,\text{Hz}$ AC standard deviation. Across rail-flagged sub-windows, the stream causally holds the last clean pre-rail scalar sample entering the fourth-order Butterworth $[0.5, 30]\,\text{Hz}$ bandpass state and masks the trailing $22$ sub-windows ($11.0\,\text{s}$) from valid indicators ($u_{t,m} = 0$).

For external unnotched or DC-coupled amplifiers ($f_s \in \{125, 1000\}\,\text{Hz}$), unit-normalized microvolt samples first undergo pre-notch gross rail gating ($>1,000\,\mu\text{V}$ zero-mean AC excursion or $>600\,\mu\text{V}$ jump), causal second-order $50\text{-Hz}$ and $60\text{-Hz}$ (plus harmonic, $Q = 30$) IIR notch filtering, post-notch AC rail ($>350\,\mu\text{V}$ AC or $>350\,\mu\text{V}$ inter-sub-window DC step) and $200(128/f_s)\,\mu\text{V}$ slew gating, causal post-rail DC baseline continuity (holding the pre-artifact scalar while absorbing post-rail DC steps), and unclipped ($c_{\text{clip}} = +\infty$) causal $[0.5, 30]\,\text{Hz}$ second-order-sections filtering. Valid unblanked sub-windows $u_{t,m} = 1$ with filtered maximum amplitude $\max_{k \in \mathcal{S}_{t,m}}|\tilde{v}[k]| \le 5.0\,\mu\text{V}$ define the lag-free raw suppression ratio:
\begin{equation}
\label{eq:rawbsr}
\text{BSR}_{\text{raw},t} = 100 \times \frac{\sum_{m=1}^{60} u_{t,m}\,\mathbb{I}\!\left(\max_{k \in \mathcal{S}_{t,m}}|\tilde{v}[k]| \le 5.0\,\mu\text{V}\right)}{\sum_{m=1}^{60} u_{t,m}},
\end{equation}
evaluated when unblanked epoch coverage satisfies $\frac{1}{60}\sum_{m=1}^{60} u_{t,m} \ge 0.50$ and monitor Signal Quality Index (SQI) $\ge 50\%$ (where present). Over a $5\text{-min}$ ($H = 10$-epoch) future horizon $\mathcal{H}_t = \{t{+}1, \dots, t{+}10\}$ at epochs $t \ge 9$ with $\text{SQI}_t \ge 50\%$, we separate three sustained cortical onset endpoints ($\ge 2$ consecutive $30\text{-s}$ epochs exceeding $5\%$ inside $\mathcal{H}_t$) across distinct $5\text{-min}$ ($s \in [t{-}9, t]$) clean histories: (i) \emph{Smoothed Pre-BS} ($\exists\, t' \in \{t{+}1, \dots, t{+}9\}$ with $\min(\text{SR}_{t'}, \text{SR}_{t'+1}) > 5\%$ given $\max_{s \in [t-9,t]}\max(\text{SR}_s, \text{BSR}_{\text{raw},s}) \le 5\%$), (ii) \emph{Lag-Free Pre-RawBS} ($\min(\tilde{r}_{t'}, \tilde{r}_{t'+1}) > 5\%$ for $\tilde{r}_u = \text{BSR}_{\text{raw},u}\,\mathbb{I}(\text{SQI}_u \ge 50\%)$ given $\max_{s \in [t-9,t]}\text{BSR}_{\text{raw},s} \le 5\%$, removing $63\text{-s}$ boxcar carryover), and (iii) \emph{Strict-Zero Pre-RawBS} (same sustained $\tilde{r}$ target given $\max_{s \in [t-9,t]}\text{BSR}_{\text{raw},s} = 0\%$).

On native $2\text{-s}$ blood pressure streams $\mathbf{p}[r] = (\text{SBP}[r], \text{DBP}[r], \text{MBP}[r])$, a genuine three-channel oscillometric cuff arrival is distinguished from held broadcast repeats and $>10\text{-s}$ telemetry dropouts via:
\begin{equation}
\label{eq:cuff}
\begin{aligned}
a_{\text{3ch}}[r] &= \mathbb{I}\!\left(\Delta t_r \le 10\,\text{s} \;\wedge\; \|\mathbf{p}[r] - \mathbf{p}[r-1]\|_\infty > \tau_{\text{cuff}}\right), \\
\tau_{\text{cuff}} &= 0.5\,\text{mmHg},
\end{aligned}
\end{equation}
contrasted against threshold-free continuous innovation recency $\Delta_{\text{innov},t} = |\text{MAP}_t - \text{MAP}_{t-1}|$ and exponentially decayed multi-horizon MAP slopes. Hemodynamic early warning (\emph{Pre-MAP $< 65\,\text{mmHg}$}) conditions on a $3\text{-min}$ ($6$-epoch, $t \ge 5$) normotensive history $\min_{s \in [t-5,t]}\text{MAP}_s \ge 65\,\text{mmHg}$, targeting $\ge 2$ consecutive epochs $< 65\,\text{mmHg}$ inside $\mathcal{H}_t$ on invasive ART recordings and physical-cuff-arrival criteria on NIBP recordings ($\ge 2$ physical arrivals $< 65\,\text{mmHg}$, minimum arrival $< 62\,\text{mmHg}$ across $\ge 2$ inflations, or $< 65\,\text{mmHg}$ when a single inflation occurs in $\mathcal{H}_t$).

\subsection{Nested cortical--hemodynamic representations and \texorpdfstring{$2\times 2$}{2x2} sequence probes}
\label{sec:models}

To test whether sub-second $128\text{-Hz}$ waveform convolutions or long-context state-space recurrences add predictive signal beyond domain-engineered cortical trough physics, we constructed a strictly nested multi-timescale feature hierarchy ($5\times$ causal rolling summaries---current epoch $t$, $2\text{-min}$ mean, $10\text{-min}$ mean, $5\text{-min}$ standard deviation, and $5\text{-min}$ difference---plus $30\text{-s}$ first differences and $5\text{-min}$ extrema; full dimensions and hyperparameters in Supplementary Table~S2):
\begin{enumerate}
    \item \textbf{De-impedanced cortical + TCI GBDT ($167$ cortical, $192$ hemodynamic features):} Combines causal sub-window voltage-trough percentiles ($P_{10}, P_{25}$), gain-normalized trough ratios (dividing trough envelopes by causal $5\text{-min}$ $75$th-percentile RMS so voltage attenuation remains invariant to skin--electrode impedance scaling), $\alpha$-spindle ($8$--$14\,\text{Hz}$) collapse fractions \cite{L42_Cartailler2019}, $\log(\alpha/\delta)$ and aperiodic spectral summaries, recursive state-space BSP \cite{L34_Chemali2013}, $\text{BSR}_{\text{raw},t}$, BIS/SR/electromyography (EMG)/SQI summaries, Orchestra propofol and remifentanil effect-site concentrations ($C_e$) and infusion rates, plus $25$ continuous hemodynamic innovation-recency descriptors for MAP forecasting, fitted via $5$-seed histogram GBDT with validation Platt scaling.
    \item \textbf{Drug-agnostic ($142$ / $167$) and waveform-only ($77$) cortical GBDT:} Dropping all $25$ TCI pump features yields the drug-agnostic TIVA model transferred zero-shot to volatile surgeries (benchmarking against an in-regimen volatile end-tidal minimum alveolar concentration [MAC] oracle trained on the $649$-case volatile training pool). Dropping all proprietary BIS/hemodynamic monitor channels retains $77$ pure raw-EEG trough, spindle, BSP, and age features for external cross-monitor evaluation.
    \item \textbf{Controlled $2\times 2$ front-end $\times$ back-end probes:} A $5$-fold subject-disjoint out-of-fold $128\text{-Hz}$ causal ConvNeXt-1D stem ($3,840$ samples/epoch, stride-$4$ downsampling $128 \to 32 \to 8\,\text{Hz}$, dilated depthwise $k{=}7$ blocks producing $32$-dimensional embeddings \cite{L_Liu2022_ConvNeXt}) and a $3$-seed decoupled two-trunk selective diagonal state-space sequence model (Chrono-SSM, log-uniform timescales $\tau \in [1, 32]\,\text{min}$ \cite{L22_Gu2022,L29_Gu2024,L39_Tallec2018}) cross-plug both front-ends with both back-ends. On the held-out validation cohort ($N = 260$), Chrono-SSM did not surpass de-impedanced GBDT at the $\Delta\text{AUPRC} \ge 0.005$, $p < 0.10$ validation retention rule ($0.2031$ vs.\ $0.1793$, $\Delta = {+}0.0237$, $p = 0.2730$), locking GBDT as the primary parsimonious architecture prior to unblinding the $525$-case locked test cohort.
\end{enumerate}

\subsection{Patient-cluster subgroup-covariate and online conformal risk control}
\label{sec:crc}

Point forecasts are paired with a $6$-head Stage-B GBDT uncertainty module predicting next-epoch ($t{+}1$, $30\text{-s}$) cortical suppression $Y_{i,t}^{\text{BSR}} = \text{SR}_{i,t+1}$ (plus $\text{BSR}_{\text{raw},i,t+1}$ in Shift-2 volatile transfer) and physical-arrival native $2\text{-s}$ pressure $Y_{i,t}^{\text{MAP}} = \text{MAP}_{i,t+1}$ via $L_1$ conditional expectations $\hat{\mu}_{i,t}$, $5$-fold out-of-fold spread $\hat{\sigma}_{i,t}$, and one-sided pinball quantiles ($\hat{q}_{0.95,i,t}^{\text{BSR}}, \hat{q}_{0.05,i,t}^{\text{MAP}}$). Using floored spreads $\tilde{\sigma}_{i,t}^{\text{BSR}} = \max(\hat{\sigma}_{i,t}^{\text{BSR}}, 0.25)$ and $\tilde{\sigma}_{i,t}^{\text{MAP}} = \max(\hat{\sigma}_{i,t}^{\text{MAP}}, 1.0)$ with clamped one-sided scores $s_{i,t}^{\text{BSR}} = \max\bigl(0, (Y_{i,t}^{\text{BSR}} - \hat{\mu}_{i,t}^{\text{BSR}})/\tilde{\sigma}_{i,t}^{\text{BSR}}\bigr)$ and $s_{i,t}^{\text{MAP}} = \max\bigl(0, (\hat{\mu}_{i,t}^{\text{MAP}} - Y_{i,t}^{\text{MAP}})/\tilde{\sigma}_{i,t}^{\text{MAP}}\bigr)$, a multiplier $q \ge 0$ issues safety bounds $\hat{U}_{i,t}^{\text{BSR}}(q) = \min(100, \hat{\mu}_{i,t}^{\text{BSR}} + q\,\tilde{\sigma}_{i,t}^{\text{BSR}})$ and $\hat{L}_{i,t}^{\text{MAP}}(q) = \max(20, \hat{\mu}_{i,t}^{\text{MAP}} - q\,\tilde{\sigma}_{i,t}^{\text{MAP}})$, incurring unweighted patient-cluster exceedance loss $L_i(q) = |\mathcal{I}_i|^{-1}\sum_{t \in \mathcal{I}_i}\mathbb{I}(s_{i,t} > q) \in [0, 1]$ over valid epochs $\mathcal{I}_i$ \cite{L23_Angelopoulos2024,L_Dunn2023}. Baseline \emph{Ungrouped CQR} calibrates an additive patient-cluster offset $\hat{d}_{\text{cqr}}$ onto $\max(\hat{\mu}_{i,t}^{\text{BSR}}, \hat{q}_{0.95,i,t}^{\text{BSR}})$ and $\min(\hat{\mu}_{i,t}^{\text{MAP}}, \hat{q}_{0.05,i,t}^{\text{MAP}})$ \cite{L_Romano2019}.

To protect high-vulnerability subgroups under covariate shift without suffering $2\times 2$ cell sparsity, let $a(i) \in \{\text{adult}, \text{geriatric}\}$ index age stratum ($\ge 65\,\text{yr}$) and $m(i) \in \{\text{ART}, \text{NIBP}\}$ index blood-pressure modality, with unweighted \emph{Marginal-Envelope CRC} defined by $\hat{q}_{a,m}^{\text{env}} = \max(\hat{q}_a^{\text{age}}, \hat{q}_m^{\text{mod}})$. Using logistic domain classifiers fitted strictly on training pools ($5$ covariates in Shift~1, $8$ drug-agnostic covariates in Shift~2), Platt-calibrated epoch odds ratios $w_{i,t}^{\text{dr}} \in [0.2, 10.0]$ define within-patient importance-weighted loss $L_i^{\text{iw}}(q) = \bigl(\sum_{t \in \mathcal{I}_i} w_{i,t}^{\text{dr}}\mathbb{I}(s_{i,t} > q)\bigr) / \bigl(\sum_{t \in \mathcal{I}_i} w_{i,t}^{\text{dr}}\bigr)$ and patient cluster weight $W_i = |\mathcal{I}_i|^{-1}\sum_{t \in \mathcal{I}_i} w_{i,t}^{\text{dr}} \in [0.2, 10.0]$ with $W_{\max} = \max(1.0, \max_{1 \le i \le n} W_i)$ \cite{L55_Tibshirani2019}.

\begin{proposition}[Patient-cluster subgroup-covariate and overlapping-group risk bounds]
\label{prop:crc}
Let calibration patients $i = 1, \dots, n$ have monotone losses $L_i^{\text{iw}}(q) \in [0, 1]$ and bounded oracle likelihood-ratio weights $W_i \le W_{\max}$.
\begin{enumerate}
    \item[(i)] (\textbf{Covariate-weighted patient-cluster CRC}) Under weighted patient-cluster exchangeability between calibration and target patients with $W_{n+1} \le W_{\max}$, selecting
    \begin{equation}
    \label{eq:cov_crc}
    \hat{q}^{\text{cw}} = \inf\!\left\{q \ge 0 : \frac{\sum_{i=1}^n W_i L_i^{\text{iw}}(q) + W_{\max}}{\sum_{i=1}^n W_i + W_{\max}} \le \alpha\right\}
    \end{equation}
    satisfies $\mathbb{E}[L_{n+1}^{\text{iw}}(\hat{q}^{\text{cw}})] \le \alpha$ (empirical plug-in odds $\hat{W}_i$ approximate this bound up to density-ratio estimation slack \cite{L31_Barber2023,L55_Tibshirani2019}).
    \item[(ii)] (\textbf{Subgroup-Covariate-Weighted envelope}) Computing Eq.~\eqref{eq:cov_crc} separately within marginal age stratum $\mathcal{G}_a = \{i : a(i)=a\}$ to obtain $\hat{q}_a^{\text{age,iw}}$ and within marginal modality stratum $\mathcal{G}_m = \{i : m(i)=m\}$ to obtain $\hat{q}_m^{\text{mod,iw}}$, the conservative envelope $\hat{q}_{a,m}^{\text{scw}} = \max\bigl(\hat{q}_a^{\text{age,iw}},\, \hat{q}_m^{\text{mod,iw}}\bigr)$ satisfies $\mathbb{E}[L_{n+1}^{\text{iw}}(\hat{q}_{a(n+1),m(n+1)}^{\text{scw}}) \mid a(n+1) = a] \le \alpha$ and $\mathbb{E}[L_{n+1}^{\text{iw}}(\hat{q}_{a(n+1),m(n+1)}^{\text{scw}}) \mid m(n+1) = m] \le \alpha$ under within-margin weighted exchangeability, without partitioning $n$ into four sparse $2\times 2$ cells.
    \item[(iii)] (\textbf{Patient-cluster Overlapping-Group CRC}) Over feature map $\Phi(i) = (1, \mathbb{I}\{\text{age}_i \ge 65\}, \mathbb{I}\{\text{NIBP}_i\})^\top \in \mathbb{R}^3$, total calibration epochs $M = \sum_{i=1}^n |\mathcal{I}_i|$, and inverse cluster weights $w_{i,t} = |\mathcal{I}_i|^{-1}$, solving the HiGHS dual linear program $\max_{\zeta \in [-\alpha, 1-\alpha]^M}\sum_{i=1}^n\sum_{t \in \mathcal{I}_i} w_{i,t} s_{i,t} \zeta_{i,t}$ subject to $\sum_{i=1}^n \Phi(i)\bar{\zeta}_i + (1 - \alpha + \varepsilon)\Phi(n{+}1) = \mathbf{0}$ ($\bar{\zeta}_i = \sum_{t \in \mathcal{I}_i} w_{i,t}\zeta_{i,t}$, $\varepsilon = 10^{-8}$, conservative $S = +\infty$ test imputation, falling back to Global Split CRC $\hat{q}_{\text{global}}$ if solver status $\ne 0$) yields primal multipliers $\hat{\beta}$ and cutoff $\hat{q}_{n+1}^{\text{og}} = \max(0, \Phi(n{+}1)^\top\hat{\beta})$ satisfying $\mathbb{E}[\Phi_j(n{+}1) L_{n+1}(\hat{q}_{n+1}^{\text{og}})] \le \alpha\,\mathbb{E}[\Phi_j(n{+}1)]$ for all $j \in \{1,2,3\}$ under stationary patient exchangeability \cite{L32_Gibbs2025}.
\end{enumerate}
\end{proposition}
The proof is given in \ref{app:proof}. Finally, to adapt online across unmodeled conditional drift without future leakage, let $\mathcal{U}_{t-d} = \{u \in \{1, \dots, t{-}d\} : Y_{i,u} \text{ observed}\}$ denote valid realized steps up to feedback lag $d$ ($d_{\text{BSR}} = 2$ epochs [$60\,\text{s}$] and $d_{\text{MAP}} = 1$ epoch [$30\,\text{s}$]). \emph{Causal Online Conformal PID} \cite{L30_Angelopoulos2023} initializes proportional state $q_{P,1} = \hat{q}_{a(i),m(i)}^{\text{env}}$ and updates strictly from realized errors $e_u = \mathbb{I}(s_{i,u} > q_u)$ ($u \in \mathcal{U}_{t-d}$):
\begin{equation}
\label{eq:pid}
\begin{aligned}
q_{P,t+1} &= \max\bigl(0,\, q_{P,t} + \eta(e_{t-d} - \alpha)\bigr), \\
q_t &= \max\!\left(0,\, q_{P,t} + k_I\,\tanh\!\left(\frac{\sum_{u \in \mathcal{U}_{t-d}}(e_u - \alpha)}{\sqrt{|\mathcal{U}_{t-d}| + 1}}\right)\right),
\end{aligned}
\end{equation}
holding $q_{P,t+1} = q_{P,t}$ when $t{-}d \notin \mathcal{U}_{t-d}$ (zero $\tanh$ term when $|\mathcal{U}_{t-d}| = 0$), with $\eta = 0.16 D$ and $k_I = 0.10 D$ scaled to calibration spread $D = \max(P_{90}(s) - P_{50}(s), 0.5)$ (Supplementary Proposition~S6.1).

\section{Experimental Results}
\label{sec:results}

\subsection{Pillar 1: Bedside hardware artifact disentangling across four cohorts}
\label{sec:res_dsp}

Across the $500$-case VitalDB TIVA signal-audit subset ($175,766$ epochs) and the $1,048$-case locked volatile cohort ($393,024$ epochs), our pre-filter Nyquist and rail detector isolated $8,628$ and $11,739$ cyclic BIS Vista impedance-check events at median inter-event intervals of $10.13\,\text{min}$ (interquartile range [IQR] $[10.09, 10.13]$ and $[10.13, 10.13]\,\text{min}$), blanking $6.02\%$ of raw sub-windows (Fig.~\ref{fig:dsp}b, Table~\ref{tab:cohort}). Without pre-filter blanking, bandpass filter recovery from cyclic test pulses created $7,013$ TIVA and $18,043$ volatile epochs with apparent $>5\%$ raw flatlines while commercial BIS-SR simultaneously reported $0\%$. Causal pre-filter blanking eliminated $97.90\%$ ($7,013 \to 147$) of TIVA and $97.60\%$ ($18,043 \to 433$) of volatile zero-SR flatline conflicts---including $93.26\%$ ($10,219 / 10,957$) of false lag-free and $95.44\%$ ($9,446 / 9,897$) of false strict-zero onset windows on volatile surgeries and $86.26\%$ ($837 \to 115$) of emergence artifacts---while retaining $99.23\%$ ($4,001 / 4,032$) and $98.96\%$ ($3,037 / 3,069$) of concordant deep burst suppression ($\text{BIS-SR} \ge 20\%$).

On external European DOSE-I endoscopy recordings ($N = 171$, $125\,\text{Hz}$), post-notch AC conditioning yielded $97.06\%$ dual-channel valid epochs with zero mains wipeouts ($99.75\%$ valid when independent benchmark artifact annotations covered $<50\%$ of an epoch vs.\ $83.05\%$ when $\ge 50\%$; causal $30$--$47\,\text{Hz}$ EMG power correlated at $r = {+}0.3107$ with external artifact burden). On the $N = 8$ DC-coupled OpenNeuro ds004541 qualitative sanity check ($3.2$--$17.4\,\text{mV}$ $|\text{DC}|$ baseline offsets), legacy pre-notch microvolt gating rejected $98.73\%$ of channel epochs ($1.27\%$ yield), whereas unclipped DC-safe post-notch conditioning recovered $74.09\%$ channel yield ($81.31\%$ fused frontal montage yield) with zero clipping-floor epochs, near-zero non-suppressed $\text{BSR}_{\text{raw}}$ ($0.05\%$ during maintenance), low TIVA suppression scores ($0.0071$--$0.0153$), and expected surgical $\alpha$--$\delta$ spectral slowing relative to post-recovery wakefulness ($\Delta\log(\alpha/\delta) = -0.8059$ $[-1.3797, -0.2862]$, $p = 0.0020$; Fig.~\ref{fig:dsp}c, Supplementary Table~S9).

Auditing $1,294,016$ native $2\text{-s}$ NIBP packets across $455$ cuff-monitored TIVA cases confirmed that $98.85\%$ of $2\text{-s}$ samples are held broadcast repeats ($83.24\%$ on DOSE-I Philips Intellivue across $1,567$ cuff inflations, median interval $2.95\,\text{min}$). Three-channel $(\text{SBP}, \text{DBP}, \text{MBP})$ gating isolated $14,019$ physical cuff inflations (median interval $2.47\,\text{min}$, IQR $[1.90, 2.57]\,\text{min}$)---capturing ${+}5.98\%$ ($839$) pulse-pressure-only updates missed by single-channel MBP differencing---from $825$ telemetry dropouts ($\Delta t > 10\,\text{s}$). However, replacing threshold-free continuous innovation recency with discrete three-channel cuff-step indicators conferred no forecasting lift (\emph{Pre-MAP $< 65\,\text{mmHg}$} AUPRC $0.3086$ vs.\ $0.3129$, $\Delta = -0.0044$, $p = 0.4040$; Supplementary Table~S6), falsifying discrete cuff-step feature engineering while motivating physical-arrival ground-truth evaluation.

\begin{figure*}[!t]
\centering
\includegraphics[width=0.90\textwidth]{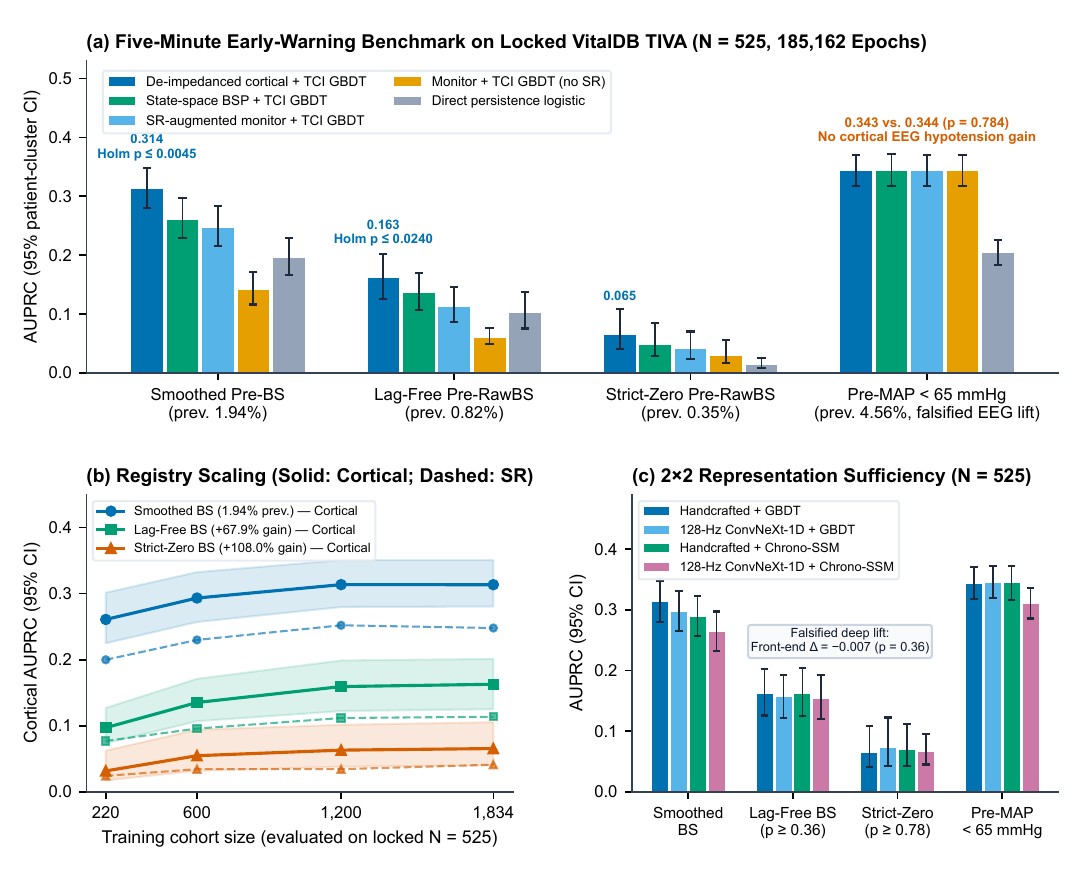}
\caption{\textbf{Full-registry five-minute early-warning benchmark, sample scaling, and $2\times 2$ representation sufficiency on locked unseen VitalDB TIVA surgeries ($N = 525$, $185,162$ epochs).} (a) Test-set AUPRC ($95\%$ patient-cluster bootstrap CIs) and $9$-comparator Holm-adjusted significance across cortical and hemodynamic endpoints. (b) Monotonic registry scaling from $N_{\text{train}} = 220$ to $1,834$ subjects (solid: de-impedanced cortical GBDT; dashed: SR-augmented clinical baseline). (c) Controlled $2\times 2$ cross-plugging showing that $128\text{-Hz}$ ConvNeXt-1D and Chrono-SSM sequence models confer no incremental lift on lag-free or strict-zero cortical suppression.}
\label{fig:tiva}
\end{figure*}

\begin{table*}[!t]
\centering
\caption{Five-minute burst-suppression and hypotension early-warning benchmark on locked unseen VitalDB TIVA surgeries ($N = 525$).}
\label{tab:tiva_benchmark}
\footnotesize
\setlength{\tabcolsep}{1.8pt}
\begin{tabular*}{\textwidth}{@{\extracolsep{\fill}}llcccccc}
\toprule
\textbf{Endpoint (prev.)} & \textbf{Model ($d$)} & \textbf{AUPRC [95\% CI]} & \textbf{AUROC [95\% CI]} & \textbf{ECE} & \textbf{$\Delta$AUPRC} & \textbf{$p_{\text{raw}}$} & \textbf{$p_{\text{Holm}}$} \\
\midrule
Smoothed BS ($1.94\%$) & \textbf{Cortical + TCI GBDT ($167$)} & $\mathbf{0.3135}$ $[0.2796, 0.3483]$ & $0.9132$ $[0.8983, 0.9268]$ & $0.0010$ & --- & --- & --- \\
 & Persistence logistic ($5$) & $0.1962$ $[0.1656, 0.2297]$ & $0.8317$ $[0.8071, 0.8539]$ & $0.0023$ & ${+}0.1173$ & $<0.0005$ & $<0.0045$ \\
 & Monitor + TCI GBDT ($65$) & $0.1412$ $[0.1161, 0.1710]$ & $0.8546$ $[0.8347, 0.8740]$ & $0.0017$ & ${+}0.1723$ & $<0.0005$ & $0.0045$ \\
 & SR-aug.\ monitor GBDT ($70$) & $0.2479$ $[0.2160, 0.2838]$ & $0.8983$ $[0.8831, 0.9130]$ & $0.0011$ & ${+}0.0656$ & $<0.0005$ & $0.0045$ \\
 & State-space BSP GBDT ($87$) & $0.2614$ $[0.2291, 0.2965]$ & $0.9009$ $[0.8859, 0.9156]$ & $0.0010$ & ${+}0.0522$ & $<0.0005$ & $0.0045$ \\
 & ConvNeXt-1D + GBDT ($193$) & $0.2972$ $[0.2647, 0.3307]$ & $0.9097$ $[0.8957, 0.9231]$ & $0.0011$ & ${+}0.0163$ & $0.0030$ & $0.0090$ \\
 & Cortical + Chrono-SSM ($35$) & $0.2889$ $[0.2565, 0.3231]$ & $0.9048$ $[0.8896, 0.9189]$ & $0.0015$ & ${+}0.0246$ & $0.0180$ & $0.0360$ \\
 & ConvNeXt-1D + Chrono-SSM ($53$) & $0.2638$ $[0.2322, 0.2972]$ & $0.9046$ $[0.8903, 0.9183]$ & $0.0010$ & ${+}0.0497$ & $<0.0005$ & $0.0045$ \\
\midrule
Lag-Free BS ($0.82\%$) & \textbf{Cortical + TCI GBDT ($167$)} & $\mathbf{0.1627}$ $[0.1257, 0.2016]$ & $0.8924$ $[0.8679, 0.9166]$ & $0.0026$ & --- & --- & --- \\
 & Persistence logistic ($5$) & $0.1034$ $[0.0754, 0.1373]$ & $0.7961$ $[0.7491, 0.8423]$ & $0.0021$ & ${+}0.0593$ & $<0.0005$ & $<0.0045$ \\
 & Monitor + TCI GBDT ($65$) & $0.0609$ $[0.0484, 0.0754]$ & $0.8458$ $[0.8134, 0.8763]$ & $0.0026$ & ${+}0.1019$ & $<0.0005$ & $0.0045$ \\
 & SR-aug.\ monitor GBDT ($70$) & $0.1135$ $[0.0868, 0.1463]$ & $0.8830$ $[0.8579, 0.9076]$ & $0.0033$ & ${+}0.0492$ & $0.0010$ & $0.0070$ \\
 & State-space BSP GBDT ($87$) & $0.1362$ $[0.1060, 0.1699]$ & $0.8885$ $[0.8622, 0.9134]$ & $0.0030$ & ${+}0.0266$ & $0.0040$ & $0.0240$ \\
 & ConvNeXt-1D + GBDT ($193$) & $0.1564$ $[0.1216, 0.1930]$ & $0.8942$ $[0.8706, 0.9177]$ & $0.0031$ & ${+}0.0063$ & $0.3860$ & $1.0000$ \\
 & Cortical + Chrono-SSM ($35$) & $0.1617$ $[0.1250, 0.2040]$ & $0.8883$ $[0.8630, 0.9139]$ & $0.0030$ & ${+}0.0011$ & $0.9460$ & $1.0000$ \\
 & ConvNeXt-1D + Chrono-SSM ($53$) & $0.1538$ $[0.1205, 0.1932]$ & $0.8857$ $[0.8611, 0.9106]$ & $0.0034$ & ${+}0.0090$ & $0.4710$ & $1.0000$ \\
\midrule
Strict-Zero BS ($0.35\%$) & \textbf{Cortical + TCI GBDT ($167$)} & $\mathbf{0.0653}$ $[0.0400, 0.1079]$ & $0.8417$ $[0.8051, 0.8801]$ & $0.0017$ & --- & --- & --- \\
 & Persistence logistic ($5$) & $0.0135$ $[0.0078, 0.0254]$ & $0.6386$ $[0.5455, 0.7324]$ & $0.0014$ & ${+}0.0518$ & $<0.0005$ & $<0.0045$ \\
 & Monitor + TCI GBDT ($65$) & $0.0301$ $[0.0175, 0.0556]$ & $0.8050$ $[0.7480, 0.8573]$ & $0.0016$ & ${+}0.0353$ & $0.0010$ & $0.0080$ \\
 & SR-aug.\ monitor GBDT ($70$) & $0.0409$ $[0.0233, 0.0702]$ & $0.8314$ $[0.7928, 0.8726]$ & $0.0015$ & ${+}0.0244$ & $0.0230$ & $0.1610$ \\
 & State-space BSP GBDT ($87$) & $0.0482$ $[0.0283, 0.0839]$ & $0.8394$ $[0.7993, 0.8826]$ & $0.0016$ & ${+}0.0171$ & $0.1170$ & $0.7020$ \\
 & ConvNeXt-1D + GBDT ($193$) & $0.0724$ $[0.0428, 0.1224]$ & $0.8362$ $[0.7942, 0.8783]$ & $0.0017$ & ${-}0.0071$ & $0.4250$ & $1.0000$ \\
 & Cortical + Chrono-SSM ($35$) & $0.0694$ $[0.0426, 0.1118]$ & $0.8301$ $[0.7847, 0.8755]$ & $0.0018$ & ${-}0.0040$ & $0.7510$ & $1.0000$ \\
\midrule
Pre-MAP $<65$ ($4.56\%$) & \textbf{Cortical + TCI GBDT ($192$)} & $\mathbf{0.3432}$ $[0.3176, 0.3705]$ & $0.8893$ $[0.8771, 0.9006]$ & $0.0029$ & --- & --- & --- \\
 & Monitor + TCI + recency ($90$) & $0.3437$ $[0.3179, 0.3707]$ & $0.8905$ $[0.8784, 0.9016]$ & $0.0029$ & ${-}0.0005$ & $0.7840$ & $1.0000$ \\
 & MAP + slope persistence ($5$) & $0.2043$ $[0.1829, 0.2262]$ & $0.8208$ $[0.8061, 0.8353]$ & $0.0104$ & ${+}0.1389$ & $<0.0005$ & $<0.0045$ \\
 & Cortical + Chrono-SSM ($28$) & $0.3453$ $[0.3166, 0.3723]$ & $0.8918$ $[0.8798, 0.9030]$ & $0.0029$ & ${-}0.0021$ & $0.7990$ & $1.0000$ \\
\bottomrule
\end{tabular*}
\par\vspace{2pt}\noindent{\footnotesize Notes: Evaluated single-shot across $N = 525$ unseen TIVA surgeries ($185,162$ total $30\text{-s}$ epochs: $145,028$ Smoothed Pre-BS, $150,742$ Lag-Free Pre-RawBS, $127,000$ Strict-Zero Pre-RawBS, and $152,066$ Pre-MAP $< 65\,\text{mmHg}$ clean-history windows). $\Delta$AUPRC denotes reference minus comparator; two-sided paired $B = 2,000$ patient-cluster bootstrap $p_{\text{raw}}$ (floored at $< 1/(B+1) \approx 0.0005$) and $9$-comparator step-down Holm--Bonferroni $p_{\text{Holm}}$ are reported alongside $10$-bin Expected Calibration Error (ECE). Complete $10$-model results across all three perioperative scopes appear in Supplementary Tables~S3--S5.}
\end{table*}

\subsection{Pillar 2: Representation sufficiency and physiological boundaries on locked TIVA (\texorpdfstring{$N = 525$}{N = 525})}
\label{sec:res_tiva}

Single-shot evaluation on the locked unseen TIVA test cohort ($N = 525$, $185,162$ epochs; Fig.~\ref{fig:tiva}, Table~\ref{tab:tiva_benchmark}; full $10$-model tables across all three perioperative scopes in Supplementary Tables~S3--S5) established consistent cortical early-warning gains alongside two sharp physiological boundaries:
\begin{enumerate}
    \item \textbf{Cortical burst-suppression early warning:} De-impedanced cortical + TCI GBDT achieved AUPRC $0.3135$ $[0.2796, 0.3483]$ (AUROC $0.9132$, $10$-bin Expected Calibration Error [ECE] $0.0010$) on \emph{Smoothed Pre-BS} ($1.94\%$ prevalence, $n_+ = 2,813$), surpassing direct persistence ($0.1962$, $p_{\text{Holm}} < 0.0045$), monitor + TCI without SR ($0.1412$, $p_{\text{Holm}} = 0.0045$), SR-augmented monitor + TCI ($0.2479$, $p_{\text{Holm}} = 0.0045$), and recursive state-space BSP + TCI ($0.2614$, $p_{\text{Holm}} = 0.0045$). When $63\text{-s}$ monitor carryover was removed on \emph{Lag-Free Pre-RawBS} ($0.82\%$ prevalence, $n_+ = 1,232$), cortical GBDT reached AUPRC $0.1627$ $[0.1257, 0.2016]$ (AUROC $0.8924$, ECE $0.0026$), outperforming persistence ($0.1034$, $p_{\text{Holm}} < 0.0045$), monitor + TCI ($0.0609$, $p_{\text{Holm}} = 0.0045$), SR-augmented GBDT ($0.1135$, $p_{\text{raw}} = 0.0010$, $p_{\text{Holm}} = 0.0070$), and state-space BSP ($0.1362$, $p_{\text{raw}} = 0.0040$, $p_{\text{Holm}} = 0.0240$). On \emph{Strict-Zero Pre-RawBS} ($0.35\%$ prevalence, $n_+ = 445$), cortical GBDT reached $0.0653$ $[0.0400, 0.1079]$ versus $0.0135$ for persistence ($p_{\text{Holm}} < 0.0045$) and $0.0301$ for monitor + TCI ($p_{\text{Holm}} = 0.0080$).
    \item \textbf{Registry sample scaling ($N_{\text{train}} = 220 \to 1,834$):} Expanding training data $8.3$-fold monotonically lifted rare-event cortical AUPRC by ${+}67.9\%$ on \emph{Lag-Free Pre-RawBS} ($0.0969 \to 0.1351 \to 0.1592 \to 0.1627$) and ${+}108.0\%$ on \emph{Strict-Zero Pre-RawBS} ($0.0314 \to 0.0545 \to 0.0630 \to 0.0653$; Fig.~\ref{fig:tiva}b, Supplementary Table~S6), with negligible sensitivity between the full $1,834$-case training set and the $1,614$ remaining training subjects outside the $220$-case signal-audit subsample ($|\Delta\text{AUPRC}| \le 0.0105$, $p \ge 0.1040$; $\le 0.0044$ across primary endpoints).
    \item \textbf{Falsified deep sub-second and cortical hypotension lift:} Across the $2\times 2$ cross-plugged matrix (Fig.~\ref{fig:tiva}c, Supplementary Table~S7), neither the $128\text{-Hz}$ ConvNeXt-1D stem ($0.1564$, $p_{\text{raw}} = 0.3860$; front-end main effect $-0.0071$, $p = 0.3600$) nor Chrono-SSM ($0.1617$, $p_{\text{raw}} = 0.9460$; back-end main effect $-0.0019$, $p = 0.8640$) improved \emph{Lag-Free} or \emph{Strict-Zero Pre-RawBS} ($p \ge 0.3600$), indicating that causal sub-window voltage-trough percentiles and $\alpha$-spindle summaries capture the observable pre-suppression signal on $128\text{-Hz}$ clinical EEG. Likewise, on \emph{Pre-MAP $< 65\,\text{mmHg}$} ($4.56\%$ prevalence, $n_+ = 6,927$), cortical + TCI GBDT ($0.3432$ $[0.3176, 0.3705]$) showed no detectable difference relative to monitor + TCI + continuous recency without raw EEG ($0.3437$ $[0.3179, 0.3707]$, $\Delta = -0.0005$, $p_{\text{raw}} = 0.7840$), confirming that $30\text{-s}$ frontal EEG adds no binary hypotension discrimination once blood pressure dynamics and drug dosing are conditioned on.
\end{enumerate}

\begin{figure*}[!t]
\centering
\includegraphics[width=0.90\textwidth]{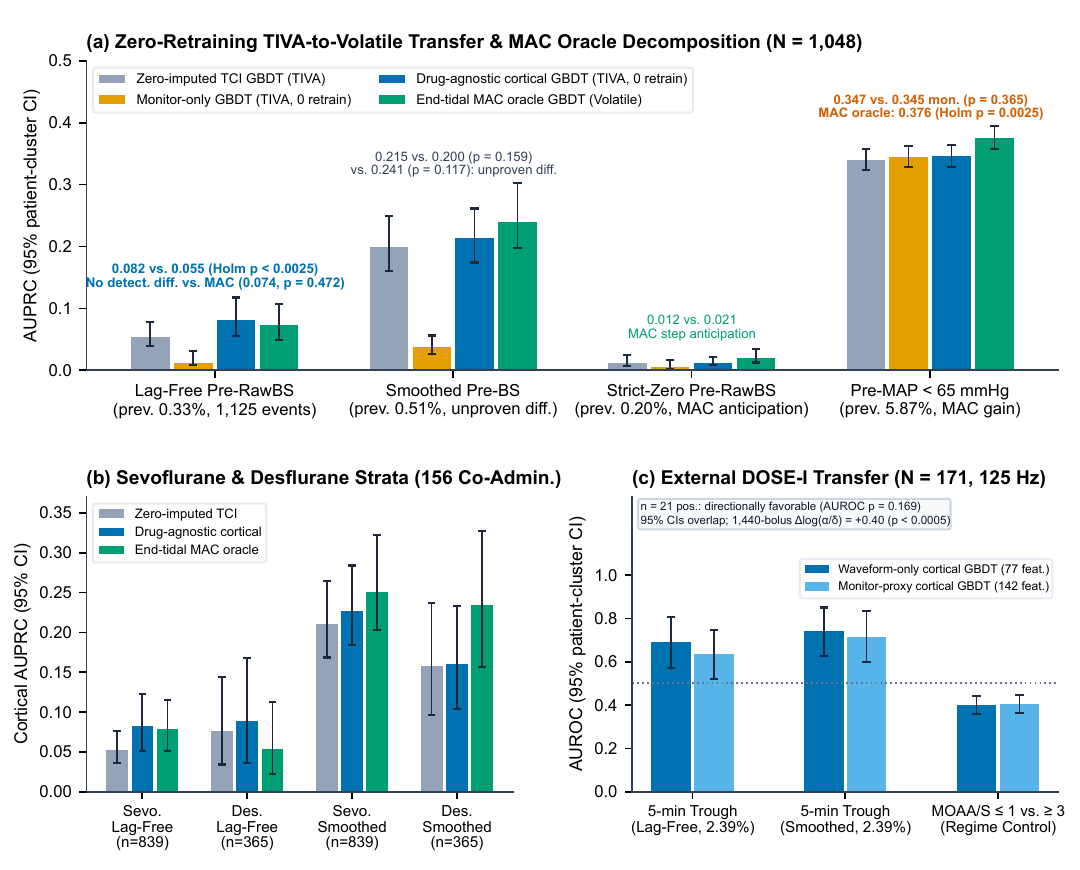}
\caption{\textbf{Zero-retraining cross-anesthetic volatile transfer ($N = 1,048$) and external European endoscopy evaluation ($N = 171$).} (a) Locked volatile cohort comparing zero-retraining TIVA-trained drug-agnostic cortical GBDT against zero-imputed TCI, clinical baselines, and an in-regimen end-tidal MAC oracle. (b) Agent-stratified sevoflurane ($n = 839$) and desflurane ($n = 365$) cortical AUPRC. (c) External DOSE-I $5\text{-min}$ trough discrimination ($n = 21$ positive patients; directionally favorable lag-free point estimate with AUROC $p = 0.1690$ and overlapping $95\%$ CIs) alongside sub-suppression MOAA/S regime-specificity control.}
\label{fig:volatile}
\end{figure*}

\begin{table*}[!t]
\centering
\caption{Zero-retraining cross-anesthetic transfer on locked volatile surgeries ($N = 1,048$) and directional external evaluation on DOSE-I ($N = 171$).}
\label{tab:volatile_external}
\footnotesize
\setlength{\tabcolsep}{1.2pt}
\begin{tabular*}{\textwidth}{@{\extracolsep{\fill}}llcccccc}
\toprule
\textbf{Cohort \& endpoint} & \textbf{Model (train, $d$)} & \textbf{AUPRC [95\% CI]} & \textbf{AUROC [95\% CI]} & \textbf{$\Delta$AUPRC} & \textbf{$p_{\text{raw}}$} & \textbf{$p_{\text{Holm}}$} & \textbf{Inference} \\
\midrule
Vol.\ Lag-Free BS ($0.33\%$) & \textbf{Drug-agnostic GBDT (TIVA, $142$)} & $\mathbf{0.0820}$ $[0.0548, 0.1176]$ & $0.8503$ $[0.8080, 0.8880]$ & --- & --- & --- & Ref.\ ($0$ retrain) \\
 & Zero-imp.\ TCI GBDT (TIVA, $167$) & $0.0547$ $[0.0389, 0.0773]$ & $0.8020$ $[0.7550, 0.8471]$ & ${+}0.0273$ & $<0.0005$ & $<0.0025$ & Significant \\
 & SR-aug.\ GBDT (TIVA, $50$) & $0.0437$ $[0.0295, 0.0742]$ & $0.8021$ $[0.7548, 0.8470]$ & ${+}0.0383$ & $<0.0005$ & $0.0025$ & Significant \\
 & Monitor-only GBDT (TIVA, $45$) & $0.0131$ $[0.0086, 0.0308]$ & $0.6958$ $[0.6422, 0.7471]$ & ${+}0.0689$ & $<0.0005$ & $0.0025$ & Significant \\
 & Persistence logistic (TIVA, $4$) & $0.0518$ $[0.0338, 0.0848]$ & $0.7426$ $[0.6951, 0.7912]$ & ${+}0.0302$ & $0.0160$ & $0.0320$ & Significant \\
 & MAC oracle GBDT (Vol., $147$) & $0.0736$ $[0.0495, 0.1074]$ & $0.8709$ $[0.8340, 0.9051]$ & ${+}0.0084$ & $0.4720$ & $0.4720$ & Unproven diff. \\
\midrule
Vol.\ Smoothed BS ($0.51\%$) & \textbf{Drug-agnostic GBDT (TIVA, $142$)} & $\mathbf{0.2147}$ $[0.1742, 0.2615]$ & $0.9429$ $[0.9289, 0.9553]$ & --- & --- & --- & Ref.\ ($0$ retrain) \\
 & Zero-imp.\ TCI GBDT (TIVA, $167$) & $0.1996$ $[0.1606, 0.2497]$ & $0.9185$ $[0.8971, 0.9379]$ & ${+}0.0150$ & $0.1590$ & $0.2340$ & Unproven diff. \\
 & SR-aug.\ GBDT (TIVA, $50$) & $0.1806$ $[0.1444, 0.2276]$ & $0.9115$ $[0.8892, 0.9318]$ & ${+}0.0340$ & $0.0010$ & $0.0040$ & Significant \\
 & Persistence logistic (TIVA, $4$) & $0.1655$ $[0.1296, 0.2118]$ & $0.8569$ $[0.8271, 0.8853]$ & ${+}0.0491$ & $0.0090$ & $0.0270$ & Significant \\
 & MAC oracle GBDT (Vol., $147$) & $0.2408$ $[0.1978, 0.3029]$ & $0.9736$ $[0.9679, 0.9793]$ & ${-}0.0262$ & $0.1170$ & $0.2340$ & Unproven diff. \\
\midrule
Vol.\ Strict-Zero ($0.20\%$) & \textbf{Drug-agnostic GBDT (TIVA, $142$)} & $\mathbf{0.0122}$ $[0.0078, 0.0208]$ & $0.7689$ $[0.7107, 0.8248]$ & --- & --- & --- & Ref.\ ($0$ retrain) \\
 & Monitor-only GBDT (TIVA, $45$) & $0.0066$ $[0.0026, 0.0169]$ & $0.6475$ $[0.5884, 0.7055]$ & ${+}0.0055$ & $0.2150$ & $0.4640$ & Unproven diff. \\
 & MAC oracle GBDT (Vol., $147$) & $0.0206$ $[0.0126, 0.0342]$ & $0.8107$ $[0.7592, 0.8608]$ & ${-}0.0084$ & $0.0600$ & $0.3000$ & Unproven diff. \\
\midrule
Vol.\ Pre-MAP $<65$ ($5.87\%$) & \textbf{Drug-agnostic GBDT (TIVA, $167$)} & $\mathbf{0.3465}$ $[0.3290, 0.3638]$ & $0.8602$ $[0.8525, 0.8671]$ & --- & --- & --- & Ref.\ ($0$ retrain) \\
 & Monitor + recency (TIVA, $70$) & $0.3452$ $[0.3279, 0.3626]$ & $0.8607$ $[0.8533, 0.8677]$ & ${+}0.0012$ & $0.3650$ & $0.3650$ & Falsified lift \\
 & MAP + slope persist.\ ($4$) & $0.2122$ $[0.1982, 0.2264]$ & $0.7940$ $[0.7842, 0.8035]$ & ${+}0.1343$ & $<0.0005$ & $<0.0025$ & Significant \\
 & MAC oracle GBDT (Vol., $172$) & $0.3758$ $[0.3572, 0.3947]$ & $0.8732$ $[0.8659, 0.8797]$ & ${-}0.0293$ & $<0.0005$ & $0.0025$ & MAC gain \\
\midrule
DOSE-I trough ($2.39\%$) & Waveform GBDT, lag-free ($77$) & $0.0510$ $[0.0204, 0.1409]$ & $0.6926$ $[0.5696, 0.8079]$ & ${+}0.0155$ & $0.0160$ & --- & Dir.\ favorable \\
 & Monitor-proxy, lag-free ($142$) & $0.0355$ $[0.0161, 0.0876]$ & $0.6335$ $[0.5210, 0.7476]$ & --- & --- & --- & Overlap CIs \\
 & Waveform GBDT, smoothed ($77$) & $0.0606$ $[0.0241, 0.1593]$ & $0.7437$ $[0.6281, 0.8503]$ & ${-}0.0036$ & $0.9250$ & --- & No det.\ diff. \\
 & Monitor-proxy, smoothed ($142$) & $0.0642$ $[0.0214, 0.1949]$ & $0.7157$ $[0.5971, 0.8331]$ & --- & --- & --- & Overlap CIs \\
\bottomrule
\end{tabular*}
\par\vspace{2pt}\noindent{\footnotesize Notes: Volatile test cohort ($N = 1,048$ sevoflurane and desflurane surgeries, $393,024$ epochs, zero propofol pump infusions) evaluates TIVA-trained models with zero retraining against an in-regimen volatile minimum alveolar concentration (MAC) oracle ($5$-comparator Holm adjustment). External European DOSE-I endoscopy cohort ($N = 171$, $n = 21$ positive patients, $3,642$ clean windows) reports paired $B = 2,000$ patient-cluster bootstrap $p_{\text{raw}}$ for $\Delta$AUPRC between waveform-only ($77$) and monitor-proxy ($142$) heads (paired AUROC $p = 0.1690$ lag-free, $p = 0.2530$ smoothed); given $n = 21$ positive patients, overlapping $95\%$ CIs, and no pre-specified equivalence margin, results are interpreted strictly as exploratory and directionally favorable (Dir.\ favorable) on lag-free onset with no detectable difference (No det.\ diff.) on smoothed onset, without superiority claims.}
\end{table*}

\subsection{Pillar 2: Cross-anesthetic volatile transfer (\texorpdfstring{$N = 1,048$}{N = 1,048}) and external DOSE-I evaluation (\texorpdfstring{$N = 171$}{N = 171})}
\label{sec:res_vol}

Transferring TIVA-trained models with zero retraining to $1,048$ locked volatile surgeries ($393,024$ epochs; Fig.~\ref{fig:volatile}a,b, Table~\ref{tab:volatile_external}; agent strata in Supplementary Table~S8) exposed structural fragility when TCI pump channels are zero-imputed (zero-imputed TCI GBDT: AUPRC $0.0547$ on \emph{Lag-Free Pre-RawBS}, $0.33\%$ prevalence, $n_+ = 1,125$). Removing TCI splits enabled the $142$-feature drug-agnostic cortical GBDT to reach AUPRC $0.0820$ $[0.0548, 0.1176]$ (AUROC $0.8503$), outperforming zero-imputed TCI (${+}0.0273$, $p_{\text{Holm}} < 0.0025$), SR-augmented ($0.0437$, $p_{\text{Holm}} = 0.0025$), monitor-only ($0.0131$, $p_{\text{Holm}} = 0.0025$), and direct persistence ($0.0518$, $p_{\text{Holm}} = 0.0320$) baselines with no detectable difference relative to the in-regimen volatile end-tidal MAC oracle ($0.0736$ $[0.0495, 0.1074]$, $p_{\text{raw}} = 0.4720$). On \emph{Smoothed Pre-BS} ($0.51\%$ prevalence), drug-agnostic GBDT ($0.2147$) outperformed SR-augmented ($0.1806$, $p_{\text{Holm}} = 0.0040$) and persistence ($0.1655$, $p_{\text{Holm}} = 0.0270$) baselines, whereas AUPRC differences versus zero-imputed TCI ($0.1996$, $p_{\text{raw}} = 0.1590$; AUROC $0.9429$ vs.\ $0.9185$, $p < 0.0005$) and the MAC oracle ($0.2408$, $p_{\text{raw}} = 0.1170$) remained unproven. Conversely, the MAC oracle achieved higher AUPRC on volatile \emph{Pre-MAP $< 65\,\text{mmHg}$} ($0.3758$ vs.\ $0.3465$, $p_{\text{Holm}} = 0.0025$; cortical vs.\ monitor-only $0.3452$, $p_{\text{raw}} = 0.3650$) and directionally higher AUPRC on rare \emph{Strict-Zero Pre-RawBS} ($0.20\%$ prevalence; $0.0206$ vs.\ $0.0122$, $p_{\text{raw}} = 0.0600$), where abrupt vaporizer step increases precede cortical voltage changes.

On external European DOSE-I propofol endoscopy sedation ($N = 171$; Fig.~\ref{fig:volatile}c, Table~\ref{tab:volatile_external}, Supplementary Table~S9), $1,440$ discrete $10\text{-mg}$ propofol boluses elicited reproducible $1$--$3\,\text{min}$ frontal $\alpha$-synchronization ($\Delta\log(\alpha/\delta) = {+}0.4012$ $[{+}0.3183, {+}0.4860]$, $p < 0.0005$) with $\text{BSR}_{\text{raw}} > 5\%$ in only $6 / 9,348$ ($0.06\%$) epochs. On $5\text{-min}$ deep-trough/suppression warning ($2.39\%$ prevalence, $87 / 3,642$ windows across $n = 21$ positive patients), the $77$-feature waveform-only cortical GBDT showed directionally favorable point discrimination on the lag-free head (no detectable difference on the smoothed head) with overlapping $95\%$ CIs relative to the $142$-feature monitor-proxy model (lag-free head AUPRC $0.0510$ vs.\ $0.0355$, paired $p_{\text{raw}} = 0.0160$; AUROC $0.6926$ $[0.5696, 0.8079]$ vs.\ $0.6335$ $[0.5210, 0.7476]$, paired $p = 0.1690$; smoothed head AUPRC $0.0606$ vs.\ $0.0642$, paired $p_{\text{raw}} = 0.9250$; AUROC $0.7437$ $[0.6281, 0.8503]$ vs.\ $0.7157$ $[0.5971, 0.8331]$, paired $p = 0.2530$). Pure burst-suppression heads appropriately yielded sub-chance discrimination on non-suppressed MOAA/S $\le 1$ versus $\ge 3$ sedation depth (AUROC $0.4011$ and $0.4049$), confirming physiological specificity to cortical suppression rather than generic hypnosis.

\begin{figure*}[!t]
\centering
\includegraphics[width=0.90\textwidth]{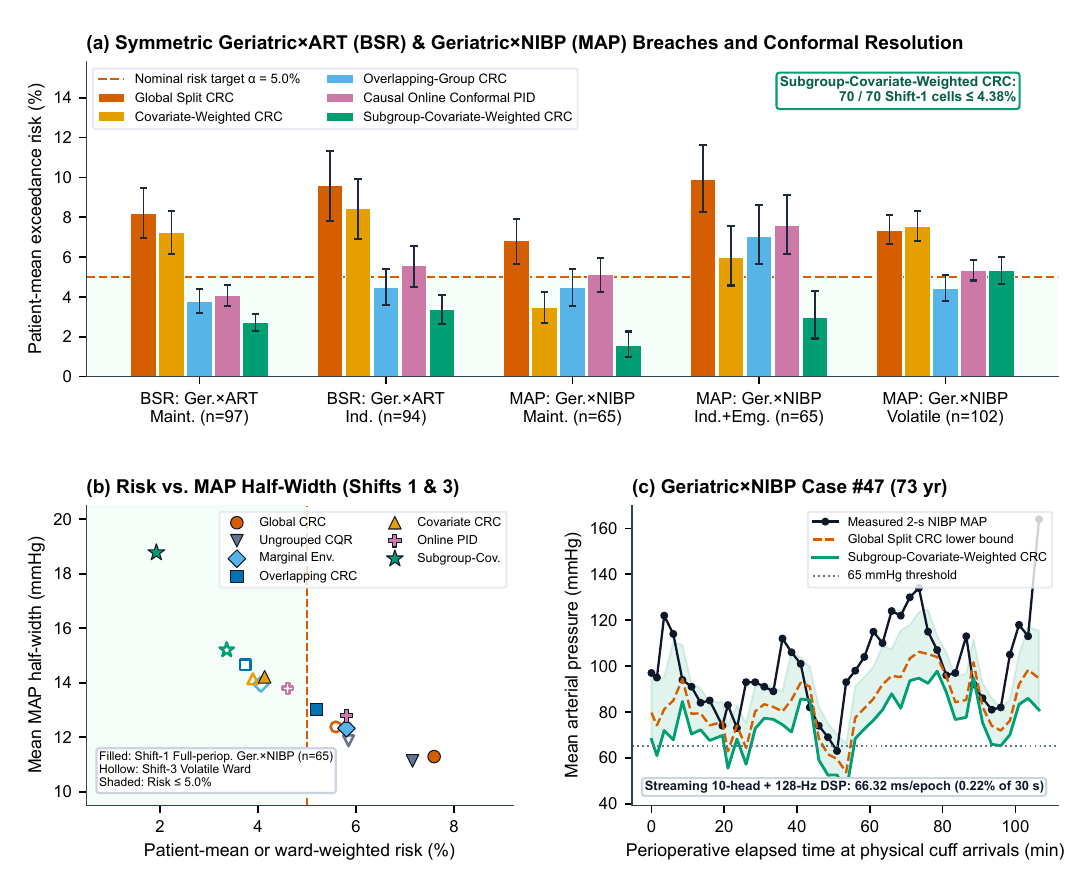}
\caption{\textbf{Three-shift conformal safety calibration ($\alpha = 5.0\%$), risk--width trade-off, and single-case bedside trajectory.} (a) Symmetric Global Split CRC subgroup exceedances on Geriatric $\times$ ART cortical BSR and Geriatric $\times$ NIBP native $2\text{-s}$ MAP across maintenance, induction/emergence, and volatile transfer, alongside resolution by Overlapping-Group CRC and Subgroup-Covariate-Weighted CRC ($70 / 70$ Shift-1 cells $\le 4.38\%$). (b) Risk versus mean MAP half-width trade-off across Shift-1 Geriatric $\times$ NIBP (filled) and Shift-3 geriatric-NIBP target ward (hollow). (c) Representative $73\text{-yr}$ Geriatric $\times$ NIBP surgery illustrating conservative protection of physical cuff hypotension excursions at $66.32\,\text{ms}$ per $30\text{-s}$ epoch.}
\label{fig:conformal}
\end{figure*}

\begin{table*}[!t]
\centering
\caption{Phase-shift and subgroup conformal risk control ($\alpha = 5.0\%$) on locked VitalDB TIVA surgeries ($N = 525$, Shift 1).}
\label{tab:shift1_conformal}
\footnotesize
\setlength{\tabcolsep}{3.0pt}
\begin{tabular*}{\textwidth}{@{\extracolsep{\fill}}lllcccc}
\toprule
\textbf{Safety bound and regime} & \textbf{Subgroup ($n$)} & \textbf{Conformal controller} & \textbf{Risk (\%)} & \textbf{95\% CI (\%)} & \textbf{P90 (\%)} & \textbf{Mean half-width} \\
\midrule
Cortical BSR upper, Maint. & Geriatric $\times$ ART ($n{=}97$) & Global Split CRC & $\mathbf{8.20}$ & $[6.96, 9.48]$ & $16.98$ & $1.02\,\text{\%}$ \\
 &  & Marginal-Envelope CRC & $4.88$ & $[4.16, 5.60]$ & $10.55$ & $1.51\,\text{\%}$ \\
 &  & Overlapping-Group CRC & $3.80$ & $[3.21, 4.38]$ & $8.23$ & $1.74\,\text{\%}$ \\
 &  & Covariate-Weighted CRC & $\mathbf{7.23}$ & $[6.14, 8.33]$ & $14.81$ & $1.14\,\text{\%}$ \\
 &  & \textbf{Subgroup-Covariate-Weighted CRC} & $2.72$ & $[2.27, 3.16]$ & $5.95$ & $2.07\,\text{\%}$ \\
 &  & Causal Online Conformal PID & $4.07$ & $[3.53, 4.61]$ & $7.52$ & $1.64\,\text{\%}$ \\
\midrule
Cortical BSR upper, Ind. & Geriatric $\times$ ART ($n{=}94$) & Global Split CRC & $\mathbf{9.59}$ & $[7.80, 11.31]$ & $24.10$ & $1.24\,\text{\%}$ \\
 &  & Marginal-Envelope CRC & $\mathbf{5.95}$ & $[4.77, 7.10]$ & $14.24$ & $1.83\,\text{\%}$ \\
 &  & Overlapping-Group CRC & $4.48$ & $[3.61, 5.38]$ & $10.47$ & $2.12\,\text{\%}$ \\
 &  & Covariate-Weighted CRC & $\mathbf{8.44}$ & $[6.89, 9.94]$ & $20.49$ & $1.39\,\text{\%}$ \\
 &  & \textbf{Subgroup-Covariate-Weighted CRC} & $3.37$ & $[2.63, 4.11]$ & $8.90$ & $2.51\,\text{\%}$ \\
 &  & Causal Online Conformal PID & $\mathbf{5.56}$ & $[4.49, 6.57]$ & $12.24$ & $1.95\,\text{\%}$ \\
\midrule
Native $2\text{{-s}}$ MAP lower, Maint. & Geriatric $\times$ NIBP ($n{=}65$) & Global Split CRC & $\mathbf{6.81}$ & $[5.66, 7.91]$ & $12.64$ & $9.85\,\text{mmHg}$ \\
 &  & Marginal-Envelope CRC & $\mathbf{5.27}$ & $[4.24, 6.28]$ & $11.23$ & $10.75\,\text{mmHg}$ \\
 &  & Overlapping-Group CRC & $4.49$ & $[3.56, 5.41]$ & $9.09$ & $11.35\,\text{mmHg}$ \\
 &  & Covariate-Weighted CRC & $3.45$ & $[2.68, 4.24]$ & $7.69$ & $12.40\,\text{mmHg}$ \\
 &  & \textbf{Subgroup-Covariate-Weighted CRC} & $1.58$ & $[0.96, 2.27]$ & $5.84$ & $16.39\,\text{mmHg}$ \\
 &  & Causal Online Conformal PID & $\mathbf{5.13}$ & $[4.25, 5.97]$ & $9.09$ & $11.20\,\text{mmHg}$ \\
\midrule
Native $2\text{{-s}}$ MAP lower, Ind.+Emg. & Geriatric $\times$ NIBP ($n{=}65$) & Global Split CRC & $\mathbf{9.91}$ & $[8.29, 11.61]$ & $17.58$ & $14.48\,\text{mmHg}$ \\
 &  & Marginal-Envelope CRC & $\mathbf{7.43}$ & $[5.89, 9.08]$ & $15.33$ & $15.81\,\text{mmHg}$ \\
 &  & Overlapping-Group CRC & $\mathbf{7.05}$ & $[5.64, 8.64]$ & $13.57$ & $16.69\,\text{mmHg}$ \\
 &  & Covariate-Weighted CRC & $\mathbf{5.97}$ & $[4.58, 7.56]$ & $13.57$ & $18.23\,\text{mmHg}$ \\
 &  & \textbf{Subgroup-Covariate-Weighted CRC} & $2.99$ & $[1.91, 4.29]$ & $9.09$ & $24.06\,\text{mmHg}$ \\
 &  & Causal Online Conformal PID & $\mathbf{7.62}$ & $[6.18, 9.13]$ & $15.33$ & $16.35\,\text{mmHg}$ \\
\midrule
Native $2\text{{-s}}$ MAP lower, Full & All patients ($n{=}524$) & Global Split CRC & $\mathbf{5.11}$ & $[4.80, 5.42]$ & $9.68$ & $8.44\,\text{mmHg}$ \\
 &  & Marginal-Envelope CRC & $4.34$ & $[4.06, 4.62]$ & $8.26$ & $9.02\,\text{mmHg}$ \\
 &  & Overlapping-Group CRC & $4.34$ & $[4.07, 4.61]$ & $8.00$ & $9.06\,\text{mmHg}$ \\
 &  & Covariate-Weighted CRC & $2.85$ & $[2.62, 3.07]$ & $5.85$ & $10.62\,\text{mmHg}$ \\
 &  & \textbf{Subgroup-Covariate-Weighted CRC} & $1.40$ & $[1.28, 1.53]$ & $3.28$ & $13.44\,\text{mmHg}$ \\
 &  & Causal Online Conformal PID & $4.95$ & $[4.74, 5.15]$ & $7.55$ & $8.87\,\text{mmHg}$ \\
\bottomrule
\end{tabular*}
\par\vspace{2pt}\noindent{\footnotesize Notes: All offline controllers are calibrated strictly on surgical maintenance of the held-out calibration cohort ($N_{\text{cal}} = 315$) at nominal risk $\alpha = 5.0\%$. Bold risk entries indicate subgroup exceedances above $5.0\%$. Subgroup-Covariate-Weighted CRC bounds all $70$ of $70$ Shift-1 target $\times$ regime $\times$ subgroup cells at $\le 4.38\%$ (complete matrix in Supplementary Tables~S10--S11).}
\end{table*}

\begin{table*}[!t]
\centering
\caption{Cross-anesthetic (Shift 2), target-ward prevalence (Shift 3) native $2\text{-s}$ MAP conformal calibration ($\alpha = 5.0\%$), and bedside streaming latency.}
\label{tab:shift_latency}
\footnotesize
\setlength{\tabcolsep}{1.8pt}
\begin{tabular*}{\textwidth}{@{\extracolsep{\fill}}llccccc}
\toprule
\textbf{Shift setting} & \textbf{Conformal controller} & \textbf{Cohort risk (\%)} & \textbf{Ger.\ NIBP (\%)} & \textbf{Half-width} & \textbf{Epoch (ms)} & \textbf{Case (ms)} \\
\midrule
Shift 2: TIVA $\to$ Volatile ($N{=}1,048$) & Global Split CRC (zero-imputed TCI) & $\mathbf{5.22}$ & $\mathbf{9.12}$ & $8.59$ & $66.32$ & $802.64$ \\
 & Global Split CRC (drug-agnostic) & $4.16$ & $\mathbf{7.37}$ & $8.76$ & $66.52$ & --- \\
 & \textbf{Overlapping-Group CRC (drug-agnostic)} & $3.74$ & $4.44$ & $9.23$ & $66.52$ & --- \\
 & Subgroup-Covariate-Weighted CRC & $3.45$ & $\mathbf{5.33}$ & $9.48$ & $66.52$ & --- \\
 & Causal Online Conformal PID & $4.61$ & $\mathbf{5.35}$ & $8.91$ & $66.52$ & --- \\
\midrule
Shift 3: Target ward on TIVA ($N{=}525$) & Global Split CRC & $\mathbf{5.21}$ & $\mathbf{5.92}$ & $10.89$ & $66.32$ & $802.64$ \\
 & Overlapping-Group CRC & $3.69$ & $4.01$ & $12.80$ & $66.32$ & $802.64$ \\
 & \textbf{Subgroup-Covariate-Weighted CRC} & $3.36$ & $4.00$ & $13.25$ & $66.32$ & $802.64$ \\
 & Causal Online Conformal PID & $4.41$ & $4.62$ & $12.14$ & $66.32$ & $802.64$ \\
\midrule
Shift 3: Target ward on Volatile ($N{=}1,048$) & Global Split CRC & $\mathbf{5.59}$ & $\mathbf{6.56}$ & $12.38$ & $66.52$ & --- \\
 & Overlapping-Group CRC & $3.75$ & $4.10$ & $14.66$ & $66.52$ & --- \\
 & \textbf{Subgroup-Covariate-Weighted CRC} & $3.36$ & $3.99$ & $15.20$ & $66.52$ & --- \\
 & Causal Online Conformal PID & $4.61$ & $4.87$ & $13.79$ & $66.52$ & --- \\
\bottomrule
\end{tabular*}
\par\vspace{2pt}\noindent{\footnotesize Notes: Shift~2 calibrates on full perioperative TIVA trajectories ($N_{\text{cal}} = 315$) and evaluates on locked volatile surgeries ($N = 1,048$; Geriatric $\times$ NIBP $n = 102$). Shift~3 calibrates on full perioperative trajectories of a $1{:}1{:}1{:}1$ balanced TIVA subsample ($N_{\text{cal}} = 156$, $39$ per cell) and evaluates under geriatric-NIBP ward weights $(0.05, 0.20, 0.15, 0.60)$. Streaming epoch latency includes causal $128\text{-Hz}$ DSP ($2.26\,\text{ms}$), feature normalization ($0.02\,\text{ms}$), $4$-endpoint $5$-seed GBDT classification ($34.29\,\text{ms}$), and $6$-head Stage-B conformal inference ($29.76\,\text{ms}$; Supplementary Table~S12).}
\end{table*}

\subsection{Pillar 3: Three-shift conformal risk control and bedside streaming latency}
\label{sec:res_crc}

Evaluating next-epoch ($t{+}1$, $30\text{-s}$) safety bounds at nominal risk $\alpha = 5.0\%$ revealed symmetric demographic-by-modality under-coverage when standard Global Split CRC was deployed across phase, drug, and ward-prevalence shifts (Fig.~\ref{fig:conformal}a, Table~\ref{tab:shift1_conformal}; complete $70$-cell matrices in Supplementary Tables~S10--S11):
\begin{enumerate}
    \item \textbf{Shift 1 (TIVA maintenance calibration $\to$ full perioperative and subgroup regimes):} Calibrating strictly on held-out TIVA surgical maintenance ($N_{\text{cal}} = 315$) and deploying on locked unseen TIVA surgeries ($N = 525$), Global Split CRC on next-epoch cortical BSR upper bounds breached Geriatric $\times$ ART ($n = 97$) at $8.20\%$ $[6.96\%, 9.48\%]$ during maintenance, $9.59\%$ $[7.80\%, 11.31\%]$ during induction ($n = 94$), and $8.01\%$ across full surgery; pure Covariate-Weighted CRC left maintenance at $7.23\%$ because age-related $\alpha$ attenuation increases conditional BSR variance beyond covariate shift alone. Conversely, on next-epoch native $2\text{-s}$ physical-arrival MAP lower bounds, Global Split CRC breached Geriatric $\times$ NIBP ($n = 65$) at $6.81\%$ $[5.66\%, 7.91\%]$ during maintenance, $9.91\%$ $[8.29\%, 11.61\%]$ across induction and emergence, and $7.60\%$ across full surgery ($5.11\%$ overall across $n = 524$ evaluable cases). Patient-cluster Overlapping-Group CRC resolved both stationary maintenance breaches simultaneously ($3.80\%$ on Geriatric $\times$ ART BSR; $4.49\%$ on Geriatric $\times$ NIBP MAP). Pairing marginal subgroup conditioning with training-fitted odds weights in Subgroup-Covariate-Weighted CRC bounded all $70$ of $70$ Shift-1 target $\times$ regime $\times$ subgroup cells at $\le 4.38\%$ ($2.72\%$ and $3.37\%$ on Geriatric $\times$ ART maintenance and induction BSR; $1.58\%$ and $2.99\%$ on Geriatric $\times$ NIBP maintenance and induction/emergence MAP; $1.40\%$ overall full-surgery MAP at $13.44\,\text{mmHg}$ mean half-width; Fig.~\ref{fig:conformal}b,c). Complementarily, Causal Online Conformal PID preserved tight overall full-surgery MAP calibration ($4.95\%$ at $8.87\,\text{mmHg}$ mean half-width) while compressing $90$th-percentile ($P_{90}$) patient risk on Geriatric $\times$ NIBP from $12.75\%$ to $8.61\%$ ($12.79\,\text{mmHg}$ half-width).
    \item \textbf{Shift 2 (full-perioperative TIVA calibration $\to$ $1,048$ volatile surgeries):} Calibrating on full perioperative TIVA trajectories ($N_{\text{cal}} = 315$) and transferring with zero retraining to volatile surgeries (Table~\ref{tab:shift_latency}), replacing zero-imputed TCI Stage-B heads (Global Split CRC $5.22\%$ overall, $9.12\%$ on Geriatric $\times$ NIBP $n = 102$) with drug-agnostic heads lowered global risk to $4.16\%$ overall ($7.37\%$ on Geriatric $\times$ NIBP). Overlapping-Group CRC restored volatile Geriatric $\times$ NIBP MAP safety to $4.44\%$ pooled ($3.83\%$ maintenance, $4.46\%$ sevoflurane, $4.00\%$ desflurane) at $9.23\,\text{mmHg}$ mean half-width.
    \item \textbf{Shift 3 ($1{:}1{:}1{:}1$ balanced calibration $\to$ geriatric-NIBP ward prevalence) and streaming runtime:} Calibrating on full perioperative trajectories of a $1{:}1{:}1{:}1$ balanced TIVA subsample ($N_{\text{cal}} = 156$, $39$ patients per cell) and deploying onto a target ward weighted $(0.05, 0.20, 0.15, 0.60)$ toward Geriatric $\times$ NIBP caused Global Split CRC to exceed $5.0\%$ on both TIVA ($5.21\%$ ward risk, $5.92\%$ Geriatric $\times$ NIBP) and volatile ($5.59\%$ ward risk, $6.56\%$ Geriatric $\times$ NIBP) cohorts, whereas Subgroup-Covariate-Weighted CRC ($3.36\%$ and $3.36\%$), Overlapping-Group CRC ($3.69\%$ and $3.75\%$), and Causal Online Conformal PID ($4.41\%$ and $4.61\%$) maintained target validity (Fig.~\ref{fig:conformal}b, Table~\ref{tab:shift_latency}). Benchmarked end-to-end on identical hardware (NVIDIA A100-SXM4-40GB host; Supplementary Table~S12), causal $128\text{-Hz}$ DSP ($2.26\,\text{ms}$), feature normalization ($0.02\,\text{ms}$), $4$-endpoint $5$-seed GBDT classification ($34.29\,\text{ms}$), and $6$-head Stage-B conformal + PID inference ($29.76\,\text{ms}$) completed a streaming $30\text{-s}$ epoch in $66.32\,\text{ms}$ mean ($P_{95} = 68.97\,\text{ms}$, $0.22\%$ of the epoch budget) and a $3.0\text{-h}$ ($360$-epoch) case in $802.64\,\text{ms}$.
\end{enumerate}

\section{Discussion}
\label{sec:discussion}

\subsection{Disentangling bedside instrumentation physics from physiological forecasting}
Our multi-cohort findings demonstrate that apparent predictive behavior on clinical waveform archives depends critically on separating monitor instrumentation artifacts from biological state transitions before fitting statistical models. Left unblanked, cyclic $10.13\text{-min}$ BIS Vista $64\text{-Hz}$ impedance checks contaminate nearly twenty-five thousand surgical epochs across the VitalDB TIVA and volatile registries with post-bandpass $<5\,\mu\text{V}$ flatlines, corrupting both training targets and evaluation denominators for lag-free suppression onset. Similarly, external DC-coupled surgical recordings (OpenNeuro ds004541) confirm that millivolt $\text{Ag}/\text{AgCl}$ electrode offsets wipe out $98.73\%$ of valid channels if amplitude rejection or hard voltage clipping precedes AC mains notch filtering and baseline-continuous highpass conditioning. Separating commercial $63\text{-s}$ BIS-SR carryover from lag-free raw suppression lowers apparent positive prevalence from $1.94\%$ to $0.82\%$ ($0.35\%$ from strict zero) on TIVA and from $0.51\%$ to $0.33\%$ ($0.20\%$ from strict zero) on volatile anesthesia, establishing realistic precision--recall baselines for prospective clinical alerting.

\subsection{Representation sufficiency and value of negative physiological boundaries}
Controlled $2\times 2$ cross-plugging across $N = 3,214$ TIVA surgeries clarifies where model complexity helps and where physiological observability saturates. Scaling training volume from $220$ to $1,834$ procedures more than doubled strict-zero suppression AUPRC (${+}108.0\%$) and lifted lag-free suppression AUPRC by ${+}67.9\%$, confirming that rare cortical transitions benefit substantially from full-registry patient diversity. At the same time, four falsified hypotheses establish crisp engineering guardrails: (i) unblanked raw EEG flatlines largely reflect cyclic monitor impedance checks rather than silent cortex; (ii) $128\text{-Hz}$ ConvNeXt-1D convolutions and Chrono-SSM state-space recurrences add no discrimination over causal sub-window voltage-trough percentiles, gain-normalized envelopes, and $\alpha$-spindle collapse statistics on lag-free suppression ($p \ge 0.3600$), justifying a transparent $66.32\text{-ms}$ CPU tree pipeline \cite{L46_Grinsztajn2022}; (iii) $30\text{-s}$ frontal EEG features confer zero incremental lift for $5\text{-min}$ binary $\text{MAP} < 65\,\text{mmHg}$ warning on either TIVA ($p = 0.7840$) or volatile ($p = 0.3650$) surgery once prior blood pressure trajectories and anesthetic dosing are known, indicating that short-horizon arterial pressure drops are governed primarily by peripheral vascular tone, surgical stimulation, and cardiodepressant drug kinetics rather than cortical oscillations; and (iv) discrete three-channel cuff-step indicators do not improve point hypotension discrimination over continuous innovation recency ($p = 0.4040$), even though isolating genuine physical cuff arrivals remains essential for honest target labeling and conformal lower-bound calibration. Cross-anesthetic transfer to $1,048$ volatile surgeries reinforces this mechanistic picture: removing TCI features enables zero-retraining cortical transfer with no detectable difference from an in-regimen end-tidal MAC oracle on lag-free suppression ($0.0820$ vs.\ $0.0736$, $p = 0.4720$), whereas end-tidal MAC retains a clear advantage when vaporizer concentration changes drive systemic vasodilation (\emph{Pre-MAP $< 65\,\text{mmHg}$}: $0.3758$ vs.\ $0.3465$).

\subsection{Complementarity of offline subgroup-covariate and online conformal control}
Evaluating conformal controllers across $70$ TIVA shift cells and $1,048$ volatile surgeries exposes a clinically important trade-off between worst-subgroup safety and interval tightness (Fig.~\ref{fig:conformal}b). Under stationary surgical maintenance or balanced-to-ward prevalence shifts, Overlapping-Group CRC corrects both cortical Geriatric $\times$ ART and hemodynamic Geriatric $\times$ NIBP under-coverage ($3.69\%$--$4.49\%$) with modest width expansion ($9.06$--$12.80\,\text{mmHg}$). When maintenance-calibrated bounds must remain valid across high-volatility induction and emergence phases without online feedback, Subgroup-Covariate-Weighted CRC provides conservative worst-subgroup protection across all $70$ TIVA cells ($\le 4.38\%$) at the cost of wider intervals ($13.44\,\text{mmHg}$ overall, $24.06\,\text{mmHg}$ during Geriatric $\times$ NIBP induction/emergence). Conversely, when bedside software ingests realized streaming measurements with causal horizon-lag buffering, Causal Online Conformal PID tracks non-stationary baseline shifts tightly ($4.95\%$ overall risk at $8.87\,\text{mmHg}$ width) while curbing extreme individual-patient failure tails ($P_{90}$ reduced from $12.75\%$ to $8.61\%$), supporting dual display of conservative subgroup envelopes during bolus induction and adaptive online bounds during prolonged maintenance.

\subsection{Limitations}
Several limitations bound our conclusions. First, both large surgical registries (VitalDB TIVA and volatile cohorts) originate from a single tertiary academic center in South Korea; although external European (DOSE-I, $N = 171$) and DC-coupled surgical (OpenNeuro ds004541, $N = 8$) datasets validate cross-monitor conditioning and bolus pharmacodynamics, DOSE-I represents sub-suppression endoscopy sedation with only $n = 21$ deep-trough patients (yielding overlapping $95\%$ CIs, lag-free AUROC $p = 0.1690$, and strictly exploratory, directionally favorable comparisons), and ds004541 provides only a small-sample qualitative conditioning sanity check without synchronized arterial pressure. Second, rare lag-free ($0.33\%$--$0.82\%$) and strict-zero ($0.20\%$--$0.35\%$) suppression onsets impose fundamental positive-predictive-value ceilings even at AUROC $0.85$--$0.89$, meaning early-warning scores should guide clinician vigilance and conformal safety margins rather than automated closed-loop drug titration. Third, discrete surgical incision, blood-loss, and vasopressor bolus timestamps are unrecorded in open VitalDB tracks, bounding unmodeled conditional shift during acute hemorrhage or emergence (Shift-2 volatile Geriatric $\times$ NIBP Subgroup-Covariate-Weighted CRC $5.33\%$ vs.\ Overlapping-Group CRC $4.44\%$). Prospective multi-center observational validation across diverse commercial EEG monitors (SedLine, Narcotrend, Entropy) and closed-loop clinical workflow studies remain necessary prior to interventional deployment.

\section{Conclusion}
\label{sec:conclusion}

Across $5,439$ public surgical and procedural sedation recordings, disentangling cyclic $64\text{-Hz}$ impedance checks, millivolt DC baseline drift, $63\text{-s}$ suppression-ratio smoothing lag, and $2\text{-s}$ oscillometric cuff holds substantially reduces false cortical flatlines while clarifying physiological observability limits. De-impedanced cortical trough and $\alpha$-spindle GBDT models show no detectable difference from $128\text{-Hz}$ convolutional and selective state-space sequence architectures on lag-free suppression warning ($p \ge 0.3600$), transfer without retraining to volatile anesthesia on cortical endpoints, and---paired with patient-cluster Subgroup-Covariate-Weighted, Overlapping-Group, and lag-buffered Online Conformal PID calibration---restore $95\%$ subgroup safety margins in $66.32\,\text{ms}$ per $30\text{-s}$ epoch.

\appendix
\section{Proof of Proposition~\ref{prop:crc}}
\label{app:proof}
\renewcommand{\theequation}{A.\arabic{equation}}
\setcounter{equation}{0}

\begin{proof}
\emph{Part (i).} For each fixed threshold $q \ge 0$, the within-patient importance-weighted exceedance loss $L_i^{\text{iw}}(q) = \bigl(\sum_{t \in \mathcal{I}_i} w_{i,t}^{\text{dr}}\mathbb{I}(s_{i,t} > q)\bigr)/\bigl(\sum_{t \in \mathcal{I}_i} w_{i,t}^{\text{dr}}\bigr)$ is right-continuous, non-increasing in $q$, and bounded in $[0, 1]$. Under weighted patient-cluster exchangeability with oracle likelihood-ratio weights satisfying $0 < W_{n+1} \le W_{\max}$ almost surely \cite{L23_Angelopoulos2024,L55_Tibshirani2019}, monotonicity of $x \mapsto \frac{A + x}{B + x}$ on $x > 0$ when $A \le B$ together with $L_{n+1}^{\text{iw}}(q) \le 1$ implies that for all $q \ge \hat{q}^{\text{cw}}$,
\begin{equation}
\label{eq:a1}
\frac{\sum_{i=1}^n W_i L_i^{\text{iw}}(q) + W_{n+1} L_{n+1}^{\text{iw}}(q)}{\sum_{i=1}^n W_i + W_{n+1}} \le \frac{\sum_{i=1}^n W_i L_i^{\text{iw}}(q) + W_{\max}}{\sum_{i=1}^n W_i + W_{\max}} \le \alpha.
\end{equation}
Conditioning on the unordered multiset of patient pairs $\{(X_i, L_i^{\text{iw}})\}_{i=1}^{n+1}$, test index $n{+}1$ is drawn with probability proportional to $W_{n+1}$, so taking expectations over Eq.~\eqref{eq:a1} at $q = \hat{q}^{\text{cw}}$ yields $\mathbb{E}[L_{n+1}^{\text{iw}}(\hat{q}^{\text{cw}})] \le \alpha$ (when training-fitted logistic odds $\hat{w}_{i,t}^{\text{dr}} \in [0.2, 10.0]$ with $\hat{W}_{\max} = \max(1.0, \max_{i \le n}\hat{W}_i)$ substitute for unknown oracle ratios, coverage holds up to total-variation density-ratio estimation slack \cite{L31_Barber2023}).

\emph{Part (ii).} Restricting Eq.~\eqref{eq:cov_crc} to calibration patients in marginal age stratum $\mathcal{G}_a = \{i \le n : a(i) = a\}$ gives $\mathbb{E}[L_{n+1}^{\text{iw}}(\hat{q}_a^{\text{age,iw}}) \mid a(n{+}1) = a] \le \alpha$ under within-margin weighted exchangeability, and analogously $\mathbb{E}[L_{n+1}^{\text{iw}}(\hat{q}_m^{\text{mod,iw}}) \mid m(n{+}1) = m] \le \alpha$ on $\mathcal{G}_m$. Because $q \mapsto L_{n+1}^{\text{iw}}(q)$ is point-wise non-increasing and $\hat{q}_{a,m}^{\text{scw}} = \max(\hat{q}_a^{\text{age,iw}}, \hat{q}_m^{\text{mod,iw}})$, it follows for every $a$ and $m$ that
\begin{equation}
\label{eq:a2}
L_{n+1}^{\text{iw}}\!\left(\hat{q}_{a(n+1),m(n+1)}^{\text{scw}}\right) \le \min\!\left\{L_{n+1}^{\text{iw}}\!\left(\hat{q}_{a(n+1)}^{\text{age,iw}}\right),\, L_{n+1}^{\text{iw}}\!\left(\hat{q}_{m(n+1)}^{\text{mod,iw}}\right)\right\},
\end{equation}
whose conditional expectations given $a(n{+}1)=a$ and given $m(n{+}1)=m$ are both bounded by $\alpha$.

\emph{Part (iii).} Assigning each epoch $t \in \mathcal{I}_i$ of patient $i \in \{1, \dots, n{+}1\}$ inverse cluster weight $w_{i,t} = |\mathcal{I}_i|^{-1}$ ensures $\sum_{t \in \mathcal{I}_i} w_{i,t} = 1$ per patient \cite{L_Dunn2023,L_LeeBarber2026}. Consider the symmetric $(n{+}1)$-patient pinball linear program of Gibbs et al.\ \cite{L32_Gibbs2025} over $\Phi(i) \in \{0, 1\}^3$ with realized scores $\{s_{i,t}\}$ and primal solution $\hat{\beta}^{(-)}$. Its dual variables $\zeta_{i,t}^{(-)} \in [-\alpha, 1 - \alpha]$ satisfy KKT stationarity $\sum_{i=1}^{n+1} \Phi_j(i)\bar{\zeta}_i^{(-)} = 0$ ($\bar{\zeta}_i^{(-)} = \sum_{t \in \mathcal{I}_i} w_{i,t}\zeta_{i,t}^{(-)}$) for each $j \in \{1,2,3\}$, while KKT complementary slackness forces $\zeta_{i,t}^{(-)} = 1 - \alpha$ whenever $s_{i,t} > \Phi(i)^\top\hat{\beta}^{(-)}$ and $\zeta_{i,t}^{(-)} \ge -\alpha$ otherwise, yielding the epoch-wise upper bound
\begin{equation}
\label{eq:a3}
\mathbb{I}\!\left(s_{i,t} > \Phi(i)^\top \hat{\beta}^{(-)}\right) \le \zeta_{i,t}^{(-)} + \alpha \implies L_i\!\left(\Phi(i)^\top \hat{\beta}^{(-)}\right) \le \bar{\zeta}_i^{(-)} + \alpha.
\end{equation}
Imputing $s_{n+1,t} = +\infty$ fixes $\bar{\zeta}_{n+1} = 1 - \alpha$ (plus numerical tie-breaker $\varepsilon = 10^{-8} > 0$), giving the stated HiGHS dual constraint $\sum_{i=1}^n \Phi(i)\bar{\zeta}_i + (1 - \alpha + \varepsilon)\Phi(n{+}1) = \mathbf{0}$ whose primal multiplier $\hat{\beta}$ satisfies $\Phi(n{+}1)^\top\hat{\beta} \ge \Phi(n{+}1)^\top\hat{\beta}^{(-)}$ by quantile monotonicity on $\Phi(n{+}1) \ge \mathbf{0}$ (on $N_{\text{cal}} = 315$ and $156$, all four $(a,m)$ cells converged with status $0$; if infeasible, the implementation falls back to Global Split CRC $\hat{q}_{\text{global}}$, which guarantees marginal validity $\mathbb{E}[L_{n+1}(\hat{q}_{\text{global}})] \le \alpha$). Since $s_{n+1,t} \ge 0$, clamping $\hat{q}_{n+1}^{\text{og}} = \max(0, \Phi(n{+}1)^\top\hat{\beta})$ leaves exceedance unchanged, and stationary patient exchangeability across $\{1, \dots, n{+}1\}$ implies $\mathbb{E}[\Phi_j(n{+}1)\bar{\zeta}_{n+1}^{(-)}] = \frac{1}{n+1}\mathbb{E}[\sum_{i=1}^{n+1}\Phi_j(i)\bar{\zeta}_i^{(-)}] = 0$, whence $\mathbb{E}[\Phi_j(n{+}1)L_{n+1}(\hat{q}_{n+1}^{\text{og}})] \le \mathbb{E}[\Phi_j(n{+}1)(\bar{\zeta}_{n+1}^{(-)} + \alpha)] = \alpha\,\mathbb{E}[\Phi_j(n{+}1)]$.
\end{proof}

\section*{CRediT authorship contribution statement}
\textbf{Zeyu Jia:} Conceptualization, Methodology, Software, Validation, Formal analysis, Investigation, Data curation, Visualization, Writing -- original draft, Writing -- review \& editing.

\section*{Declaration of competing interest}
The author declares that he has no known competing financial interests or personal relationships that could have appeared to influence the work reported in this paper.

\section*{Funding}
This research did not receive any specific grant from funding agencies in the public, commercial, or not-for-profit sectors.

\section*{Ethics statement}
This study exclusively analyzed de-identified, publicly available human physiological recordings and involved no new human subject recruitment or prospective clinical intervention. The VitalDB surgical database was approved by the Institutional Review Board of Seoul National University Hospital (H-1408-101-605, registered at ClinicalTrials.gov NCT02914444) with a waiver of written informed consent for de-identified secondary research \cite{L27_Lee2022}. The European DOSE-I procedural sedation dataset (Zenodo \texttt{10.5281/zenodo.22145442}, CC-BY-4.0) \cite{L53_Garbe2026_DOSE1} and the OpenNeuro ds004541 general anesthesia dataset (\texttt{10.18112/openneuro.ds004541.v1.0.0}, CC0) \cite{L54_SainiFerron2023_ds004541} were collected under local ethics committee approvals and written informed consent as documented by their respective original investigators.

\section*{Data and code availability}
All four clinical datasets analyzed in this work are publicly accessible via VitalDB (\url{https://vitaldb.net/}), Zenodo (\url{https://doi.org/10.5281/zenodo.22145442}), and OpenNeuro (\url{https://doi.org/10.18112/openneuro.ds004541.v1.0.0}). Deterministic cohort manifests, streaming signal-conditioning modules, model training and evaluation scripts, and conformal risk calibration software are included in the open-source companion repository and supplementary materials.

\section*{Acknowledgments}
The author thanks the clinical investigators of Seoul National University Hospital (VitalDB), Martin Luther University Halle-Wittenberg (DOSE-I), and Pontificia Universidad Cat\'olica de Chile (OpenNeuro ds004541) for making high-resolution perioperative waveform repositories publicly available to the biomedical research community.

\bibliographystyle{elsarticle-num}
\bibliography{references}

\end{document}